\documentclass[10pt,journal,compsoc]{IEEEtran}
\usepackage{paralist}

\ifCLASSOPTIONcompsoc
  \usepackage[nocompress]{cite}
\else
  \usepackage{cite}
\fi
\ifCLASSINFOpdf
\else
\fi
\usepackage{graphicx}
\usepackage{float}
\usepackage{subfig}
\usepackage{overpic}
\usepackage[hidelinks]{hyperref}
\usepackage{url}
\usepackage{amsmath}
\usepackage{amssymb}
\usepackage{wrapfig}
\usepackage{balance}
\usepackage{color}
\usepackage{paralist}
\usepackage{booktabs}
\usepackage{tabularx}
\usepackage{svg}
\usepackage{ulem,xpatch}
\usepackage{ragged2e}

\newcommand{\change}{\textcolor{black}}

\newcommand{\modify}{\textcolor{black}}
\newcommand{\modifya}{\textcolor{black}}

\begin{document}
%
\title{\change{Towards Better Modeling with Missing Data: A Contrastive Learning-based Visual Analytics Perspective}}
%
%
%
%

\author{Laixin Xie, Yang Ouyang, Longfei Chen, Ziming Wu, and Quan Li
\IEEEcompsocitemizethanks{\IEEEcompsocthanksitem L. Xie, Y. Ouyang, L. Chen and Q. Li are with School of Information Science and Technology, ShanghaiTech University, and Shanghai Engineering Research Center of Intelligent Vision and Imaging, China. Quan Li is the corresponding author. E-mail: xielx,ouyy,chenlf,liquan@shanghaitech.edu.cn.
\IEEEcompsocthanksitem Z. Wu is with Tencent Inc., Shenzhen, Guangdong, China. E-mail: zwual@connect.ust.hk.}
\thanks{Manuscript received April 19, 2005; revised August 26, 2015.}}

%
%

\markboth{Journal of \LaTeX\ Class Files,~Vol.~14, No.~8, August~2015}%
{Shell \MakeLowercase{\textit{et al.}}: Bare Demo of IEEEtran.cls for Computer Society Journals}
%



\IEEEtitleabstractindextext{%
\begin{abstract}
\justifying{Missing data can pose a challenge for machine learning (ML) modeling. To address this, current approaches are categorized into feature imputation and label prediction and are primarily focused on handling missing data to enhance ML performance. These approaches rely on the observed data to estimate the missing values and therefore encounter three main shortcomings in imputation, including the need for different imputation methods for various missing data mechanisms, heavy dependence on the assumption of data distribution, and potential introduction of bias. This study proposes a Contrastive Learning (CL) framework to model observed data with missing values, where the ML model learns the similarity between an incomplete sample and its complete counterpart and the dissimilarity between other samples. Our proposed approach demonstrates the advantages of CL without requiring any imputation. To enhance interpretability, we introduce \textit{CIVis}, a visual analytics system that incorporates interpretable techniques to visualize the learning process and diagnose the model status. Users can leverage their domain knowledge through interactive sampling to identify negative and positive pairs in CL. The output of \textit{CIVis} is an optimized model that takes specified features and predicts downstream tasks. We provide two usage scenarios in regression and classification tasks and conduct quantitative experiments, expert interviews, and a qualitative user study to demonstrate the effectiveness of our approach. In short, this study offers a valuable contribution to addressing the challenges associated with ML modeling in the presence of missing data by providing a practical solution that achieves high predictive accuracy and model interpretability.}
\end{abstract}

}

\maketitle

\IEEEdisplaynontitleabstractindextext

%
\IEEEpeerreviewmaketitle

\IEEEraisesectionheading{\section{Introduction}\label{sec:introduction}}

\par \IEEEPARstart{M}{issing} \change{data indicates that the values in a dataset are not recorded due to a variety of factors such as inherent characteristics, privacy concerns, and difficulties in data collection. The missing data issue is prevalent in the field of machine learning (ML) and can be observed during the training and inference phases, which makes the purpose of effective modeling of the observed data challenging. If the missing values, denoted as ``NaN'', are included in the input of the model, then the ML program should throw an error. Also, the exclusion of missing values may lead to weak statistical conclusions due to a reduction in sample size~\cite{rubin2004multiple}. Consequently, effective resolution of problems caused by missing data is essential for modeling observed data with missing values.}

\par
\begin{figure}[h]
    \centering
    \includegraphics[width=0.8\linewidth]{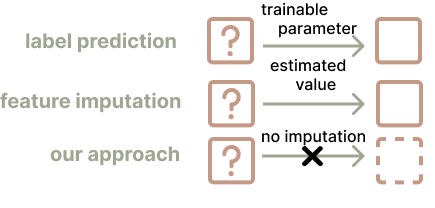}
\vspace{-3mm}
    \caption{\modify{Existing methods (i.e., label prediction and feature imputation) impute the missing data, while our approach directly utilizes the incomplete data with missing values.}}
    \vspace{-6mm}
    \label{fig:existing}
\end{figure}

 In order to obtain better performance in ML modeling, existing methods focus mainly on dealing with missing data and they are classified into two categories: \textit{feature imputation}~\cite{van2011mice,honaker2011amelia,hastie2015matrix} and \textit{label prediction}~\cite{xia2017adjusted,smieja2018processing,you2020handling}. Feature imputation fills in missing values based on the observed data distribution (\autoref{fig:existing}), \change{such as joint modeling with expectation-maximization~\cite{honaker2011amelia},} multiple imputations via chained equations (MICE)~\cite{van2011mice}, and matrix complement~\cite{hastie2015matrix}. However, the diverse missing mechanisms, missing proportions, and distributions of missing data~\cite{sarma2022evaluating} require different imputation models. In addition, assumptions about the data distribution may severely affect the imputation accuracy and introduce specific biases. Meanwhile, label prediction accomplishes the downstream ML task directly from the observed data with missing values. Particularly, the missing values are imputed by trainable parameters. Specifically, label prediction is an end-to-end strategy that adjusts these parameters by feeding only data with missing values in the training or inference phase and optimizing the prediction accuracy for downstream tasks (\autoref{fig:existing}). For example, the missing values are filled by the parametric density~\cite{smieja2018processing}, or by the intermediate prediction results~\cite{you2020handling}. In other words, most existing label prediction methods include a trainable data imputation component in their predictions, which also introduces the aforementioned disadvantage of induced bias. To summarize, the existing methods are more or less imputing missing values and inevitably produce biases (\autoref{fig:existing}).

\par Recent advances in Contrastive Learning (CL) \change{have shown} impressive performance in various \change{domains}~\cite{liu2021self}, which collects an anchor (i.e., a reference point) sample with its positive samples and disperses it with the negative samples in the latent space. In this study, CL motivates us to address \change{the issue of modeling observed data with missing values} from a CL perspective: \textit{for an incomplete sample, the ML model should learn the similarity between its complete counterpart and the dissimilarity between any other samples.} In particular, we formulate \change{the ML modeling with missing data} as a subproblem of CL and propose a CL-based framework to address the above-mentioned challenges. Our approach does not require data imputation. Instead, our approach \modify{introduces} expert knowledge in both positive and negative strategies. Moreover, \change{in contrast with} label prediction approaches, \change{our} proposed framework integrates human expertise as a guide and co-adaption of humans and systems to facilitate knowledge generation~\cite{gillies2016human}. Specifically, the framework consists of a full model and a semi-model, trained from different sets of the original dataset. The complete data records (i.e., data without missing features) are fed into the full model, \change{and the full model should have a higher numerical performance than the semi-model}. The semi-model receives incomplete records (i.e., data with specified features) and is trained to be consistent with the full model. In the inference phase, the semi-model works alone to make predictions in downstream tasks.

\par However, transferring the idea of CL to the case of missing data is nontrivial due to the following challenges. \change{First, CL-based solutions do not necessarily converge to global optimization, and the training phase may be trapped in local optimization. \change{Accordingly}, measuring and diagnosing the training phase from different perspectives are concerns that previous work has rarely attempted to address.} Second, when selecting positive and negative samples, they do not necessarily have the same dimensionality, \change{posing} a challenge when we want to \change{fully utilize} these incomplete data records. Considerable interactive modifications are required when using the sampling strategy~\cite{schroff2015facenet,suh2019stochastic,robinson2020contrastive,he2020momentum,li2021model} in our scenarios. To address the issues \change{encountered} when adapting CL concepts in our scenario, we further propose \textbf{\underline{C}}ontrastive \textbf{\underline{I}}mputation \textbf{\underline{Vis}}ualization (\textit{CIVis}), \change{a visual analytics system that facilitates} the exploration, analysis, and development of \change{CL-based modeling with missing data in a user-friendly and transparent manner.} Specifically, to address the first challenge, we derive a series of interpretable techniques~\cite{selvaraju2017grad,xuan2020improved} to measure and diagnose model status, enabling users to choose an appropriate sampling strategy \change{and} visualize the root cause of model collapse\footnote{A typical local optimization in CL, in which the deep learning model generates the same embedding regardless of the fed instance.}~\cite{grill2020bootstrap} and the improvement/degradation of model performance. \textit{CIVis} adapts several well-established sampling strategies in selecting positive and negative samples \change{to address the second challenge}. In negative sampling, users can apply various sampling strategies on the training set and \change{observe} the changes in its distributions. \change{Meanwhile,} in positive sampling, users can interactively determine the mapping relationships between incomplete and complete data. Our main contributions are as follows.
\begin{compactitem}
    \item We present a novel CL-based framework that allows model prediction with incomplete data, which does not require data imputation and can be easily generalized to different ML models.
    \item \change{We develop \textit{CIVis}, a visual analytics system, based on the \change{proposed} CL-based framework and new visualization capabilities, to support users in understanding, analyzing, and improving ML modeling of observed data with missing values.}
    \item We confirm the efficacy of our approach \modify{through} two usage scenarios with different downstream tasks (i.e., value prediction of house prices and classification of credit card repayments), quantitative experiments, expert interviews, and a qualitative user study.
\end{compactitem}

\section{Related Work}

\subsection{Methods Dealing with Missing Data}
\label{sec:missing}
\par In order to obtain better performance in ML modeling, existing approaches mainly focus on dealing with missing data and \change{address} it in two ways: \textit{feature imputation} and \textit{label prediction}. Specifically, \change{the statistics-based feature imputation}~\cite{kim2004reuse,van2011mice,honaker2011amelia,hastie2015matrix} is useful when the dataset satisfies the algorithm \change{assumptions}. For example, joint modeling~\cite{honaker2011amelia} is \change{used to populate multiple imputed values with different parameters under a specific probability density function.} Matrix completion~\cite{hastie2015matrix} requires that the matrix satisfies certain otherwise mathematically unsolvable properties. MICE~\cite{van2011mice} and KNN~\cite{kim2004reuse} are subject to the assumption of data distribution. \change{Meanwhile, deep learning-based feature imputation~\cite{gondara2017multiple,yoon2018gain} requires that the missing values fit the distribution of the observed data. Therefore, these algorithms still fill a value according to their assumptions \change{when the missing values are not within a certain assumption}, thus biasing the dataset.}

\par To overcome the limitations of feature imputation, label prediction methods~\cite{xia2017adjusted,smieja2018processing,you2020handling} predict the outcome of downstream tasks with incomplete data. The tree-based model~\cite{xia2017adjusted} is heuristic but time-consuming when the data volume is large. You et al.~\cite{you2020handling} explicitly captured the relationship between data records and features \change{by Graph Neural Network}, \change{wherein} the label predictor is based on the imputed data. Smieja et al.~\cite{smieja2018processing} imputed the corresponding positions in the feature map and predicted the labels \change{by using a neural network}. \change{These imputed values and the neural network are trained to improve the prediction \change{accuracy}.} \change{Nonetheless, imputation becomes highly challenging to diagnose when the imputed data are unavailable.} Instead of imputation, we formulate \change{modeling observed data with missing values} as a subproblem of CL \change{in this study} and allow for label prediction after removing dimensions that may \change{contain} missing data. \change{In contrast with the black box \modify{nature} of the existing label prediction methods}, our proposed CL framework is transparent. Based on our proposed CL framework, experts can interactively choose a sampling strategy that is consistent with their domain knowledge.

\subsection{Contrastive Learning}
\par CL is a popular method in self-supervised learning~\cite{he2020momentum,chen2020simple,zhu2021improving,yao2021g,zhou2021contrastive}. \change{This method is used to} relabel data and its augmentation (e.g., cropped image, flipped image) as positive and negative samples to fully utilize the data. CL first demonstrates impressive performance in the field of computer vision (e.g., classification)~\cite{he2020momentum,chen2020simple}. Some work followed the \change{loss of InfoNCE~\cite{oord2018representation}} and proposed new positive and negative sampling strategies to extend CL to other domains~\cite{zhu2021improving,yao2021g,zhou2021contrastive}. Li et al.~\cite{li2021model} applied CL to horizontal federated learning, \change{wherein} the global and local models of the last epoch generate positive and negative samples, respectively. Yao et al.~\cite{yao2021g} discussed CL in object detection, \change{wherein the} samples take the form of \change{Region of Interest (RoI)}, and the highly overlapped RoIs are negative samples of each other. In addition, negative mining has \change{long} been studied in Deep Metric Learning~\cite{schroff2015facenet,suh2019stochastic,robinson2020contrastive}\change{, with the aim of finding }similar samples from different classes, denoted as hard negative samples. Suh et al.~\cite{suh2019stochastic} defined an anchor for each class, and \change{the} classes with close anchor points are neighbors and are negative samples of each other. Schroff et al.~\cite{schroff2015facenet} defined semi-hard samples \change{as one that is slightly less distant from a positive sample than from its negative counterpart.} \change{In contrast with} the above-mentioned work that treats a domain or a particular sampling strategy as a \change{case-by-case} study whose validity and choice of sampling strategy depend on the particular usage scenario, \textit{CIVis} provides a comprehensive analysis of \change{the} different sampling strategies from the perspective of model performance and data distribution. Users can visualize the direct \change{effect} (i.e., data changes) and consequences (i.e., model changes) when choosing a specific sampling strategy.

\subsection{Visualization for Machine Learning}
\label{sec:explain}
\par The substantial work of visualization enhances the different stages of ML. A branch focuses on the assessment and correction of labeled data to improve the \modify{quality of data} used to train ML models~\cite{bors2018visual,xiang2019interactive,kandel2012profiler,angelini2022visual,sarma2022evaluating}. \change{The models} combine multiple metrics, including those for missing data, to assess \change{the} data quality, identify anomalies, and correct labels. Our work focuses on the use of missing data, rather than on assessment or correction.
\par \change{Post hoc interpretable visualizations~\cite{liu2016towards,ming2017understanding,derose2020attention,jin2022gnnlens,gou2020vatld,la2023state} are \change{regarded as} another key technique for understanding the ML black-box. \change{This mechanism is used to} build bridges between the input data and the hidden layers of a particular model to reveal how high-level information is captured layer by layer. Our work is the first to provide post hoc interpretation capabilities for CL. \change{Several} visual analytics approaches \change{have been} proposed to develop novel ML models in a progressive manner~\cite{liu2017visual,kahng2018gan,wang2018dqnviz,turkay2018progressive,echihabi2022pros}, \change{reducing} the manual effort in iterative activities. Particularly, \change{these models} diagnose the training process, explain the model status, and refine unsatisfactory performance. \change{The} progressive pipeline of our visual analytics system starts with several epochs of training, diagnoses the model with visual cues, refines the sampling strategy, and proceeds to another round of model training.}
\par \change{Some studies~\cite{zhu2021improving,zhang2022does} explain CL from a theoretical perspective. Zhu et al.~\cite{zhu2021improving} provided visual evidence (i.e. statistical plots) to diagnose the performance of CL models and perform targeted feature transformations to fix errors. Inspired by~\cite{zhu2021improving}, \textit{CIVis} also visualizes statistics, such as variance and mean of positive/negative sample distances, which \modify{facilitate} model diagnosis during the training stage.}

\begin{figure*}[h]
    \centering
    \includegraphics[width=\linewidth]{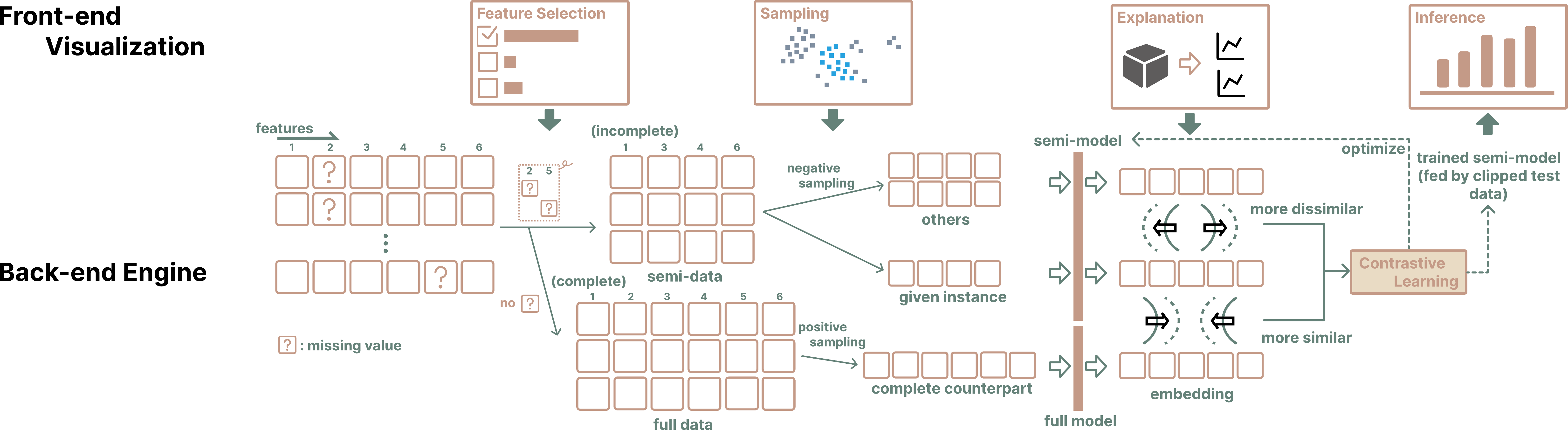}
\vspace{-6mm}
    \caption{The system pipeline. \textit{CIVis} includes a back-end engine and a front-end visualization. The back-end engine serves our proposed contrastive framework, based on a motivation that \textit{for an incomplete sample, the ML model should learn the similarity between its complete counterpart and the dissimilarity between any other samples}. The front-end visualization consists of four components (i.e., feature selection, sampling, explanation, and inference) \change{and interacts with the back-end engine}.}
    \label{fig:pipeline}
    \vspace{-6mm}
\end{figure*}

\section{Observational Study}
\subsection{Experts' Conventional Practice and Bottleneck}
\par To understand how (well) \change{the issue of modeling observed data with missing values} are tackled in practice, we worked with a team of experts from a collaborating local AI provider institution, including a product manager (E1, male, $34$), and two ML practitioners (E2, male, $28$, E3, male, $27$). A considerable part of their work involves providing Big Data and AI solutions to their clients for their specific business requirements. \change{These experts} shared with us a recent case \change{that} they encountered where they designed and developed a precision marketing strategy in a real estate scenario. Specifically, the real estate enterprise wanted to use its data to train an ML model to predict whether a particular customer would come and visit their real estate sales office and to understand the customer's characteristics. The real estate enterprise can find more suitable customers who are likely to visit the sales office from their large customer pool \change{by using this trained ML model}.

\par When utilizing ML in this context, \change{the ML practitioners (E2 and E3)} first identified training samples from the pool, \change{namely,} positive samples (i.e., those customers who have visited the sales office) and negative samples (i.e., those customers showing no interest in visiting the sales office). \change{Although} the total number of available training samples had reached approximately $25,000$, not all of the samples were good enough for training purposes. Most of the records were collected from field sales representatives and described a rough picture of customer characteristics, such as \textit{gender}, \textit{age}, \textit{occupation}, \textit{income level}, \textit{marital status}, and \textit{family structure}. E1 commented that ``\textit{if we exclude these missing records from our modeling, it may lead to weak statistical conclusions}''. \change{Specifically,} removing a large number of data records with missing characteristics may not adequately represent the population as a whole. E1 further noted that ``\textit{in the surveys we use for real estate sales offices, non-response to certain items does not occur randomly; only those with certain characteristics refuse to respond to specific questionnaires}''. Removing these missing records may bias the model training, and \change{the remaining records} may not be representative of the entire population.

\par \change{In} discussing how to \change{fill in} more records for model training, \change{E2 and E3} attempted data imputation to fill in these missing values. However, a consequence is that \change{the team} had no certainty about the \change{model} performance. Specifically, the missing data \change{are present} in the training and inference samples. During the training phase, \change{E2 and E3} performed specific data imputation strategies, such as averaging and regression, to train a good enough ML model. In the inference phase, \change{the team} must also discover any missing data in the inference samples and ensure that the feature space is the same as the training samples to ensure \change{a} smooth operation of the ML model. Although various data imputation methods \change{exist}, \change{E2 and E3 cannot easily} choose an optimal imputation model because it requires a specific level of \change{business knowledge}. Moreover, the missing data of different variables jointly determine which imputation model to use, and the use of different imputation models largely affects the quality of the data after data imputation. Therefore, \change{the team was envisioning} a more flexible way to deal with missing data rather than data imputation.

\subsection{Experts' Needs and Expectations}
\par We discussed the possibility of dealing with \change{the issue of modeling observed data with missing values} without any form of data imputation. We interviewed experts (E1–E3) to identify their primary needs and concerns. At the end of the interviews, the need to fully exploit the dataset with missing values and steer a more transparent and interactive model that incorporates domain expertise emerged as a key theme in the feedback collected. We summarize their specific requirements below.

\par \textbf{R.1 Understand the overall missing features.} Prior label prediction methods have neglected the issue of missing rates and the importance of features. E2 argued that the absence of critical features can render even human judgments unreliable. Hence, it is crucial for domain experts to possess a comprehensive understanding of the number and location of missing features, in addition to discerning the importance of each feature. Such insights enable experts to make informed decisions on how to utilize features and skip records that contain substantial amounts of missing data.

\par \textbf{R.2 Support interactive model configuration.} Domain knowledge using label prediction end-to-end models \change{typically} takes the form of strict assumptions that are difficult to modify or change. E3 commented that ``\textit{analyzing the imputation performance of label prediction is more challenging}'' because he did not know how to integrate it with his domain knowledge. Therefore, \change{the experts} wanted to represent and integrate their domain knowledge by interacting with the model in a way that would promote its interpretability and further increase their confidence in the prediction results.

\par \textbf{R.3 Evaluate model performance during training.} A subsequent response of experts is to evaluate the performance of the model and understand how much knowledge has been learned from the data and how many features have been captured. However, a potential risk \change{is} that the induced knowledge may cause the model to \change{crash or that the model does not converge well (i.e., model collapse).} Evaluation from multiple aspects can facilitate the diagnosis of the training phase and how domain knowledge is induced. \change{According to E2,} ``\textit{when a model performs well in the training phase, we have more confidence in its predictive power in the inference phase}''.

\par \textbf{R.4 Inspect the quality of prediction results.} In the inference stage, \change{the} missing features are quite unpredictable and do not provide ground truth. \change{The primary concern of experts is to assess} the quality of prediction results based on \change{the} missing data. \change{The experts} need additional auxiliary information to decide whether to accept the prediction results. \change{According to E3,} ``\textit{we will consider the missing features, the model prediction, and its interpretation together to make the final decision}''. Therefore, checking the quality of the prediction results and assessing their credibility \change{are} crucial in such a co-adaptive human-computer decision-making process.

\section{Approach Overview}
\par To meet the \change{above-mentioned} requirements, we propose a CL-based visual analytics system, \textit{CIVis}, \change{to support experts in understanding, analyzing, and improving \change{modeling observed data with missing values}.} \autoref{fig:pipeline} describes the pipeline of our approach. Specifically, \change{the experts first determine the model and features based on their missing data rates.} Accordingly, \textit{CIVis} automatically divides the data into two subsets (i.e., semi-data and full data) with different feature dimensions (i.e., complete and incomplete). \change{The data} are fed to two models with the same architecture for pre-training, which are named ``full model'' and ``semi-model'', respectively. Subsequently, the experts identify the positive and negative samples in an interactive manner in the positive sampling \change{(\autoref{positive sampling view})} and negative sampling views \change{(\autoref{negative sampling view})}. Then, the selected positive and negative samples and the full model are fed into our CL-based framework to train the semi-model. During the training phase, multiple criteria are presented in the training and comparison views \change{(\autoref{training comparison view})} to facilitate checking embedding differences and diagnosing potential model collapse. In addition, \textit{CIVis} allows users to refine their sampling strategy by considering the performance of all metrics together to obtain a sufficiently satisfactory model. In the inference phase, the data \change{are} clipped and fed into a trained semi-model for prediction. To ensure that the results can be trusted, \textit{CIVis} employs a well-established interpretability technique, \textit{GradCAM}~\cite{selvaraju2017grad}, to calculate the contribution of features to the prediction results, as shown in the inference view \change{(\autoref{inference view})}. Thus, experts can use their knowledge and the predictive capability of the model to achieve co-adaptive decisions on the final prediction outcome \change{for any incoming data (either complete or incomplete) without ground truth.}

\section{Back-end Model}
\par \modify{To mitigate the influence of imputation bias, we present a novel approach in which the task of modeling observed data with missing values is formulated as a subproblem of CL. Our proposed framework is grounded in the principles of CL and is specifically designed to address this issue.}

\subsection{Background of CL}
\par CL has a key design principle: aligning positive samples and maintaining uniformity of data distribution~\cite{wang2020understanding}. Specifically, \textbf{alignment} represents the similarity between a record and its positive samples, while \textbf{uniformity} favors a uniform distribution of record embeddings, which depends on the negative samples. For example, InfoNCE~\cite{he2020momentum} is a mathematical reflection of the design principles and the underlying CL loss function, which is described below.
\begin{equation}
    L_{con}=-log\frac{exp(\sigma(q,k_{+})/T)}{exp(\sigma(q,k_{+})/T)+\sum_{i=1}^K exp(\sigma(q,k_i)/T)},
\label{eq:con}
\end{equation}
where $\sigma$ denotes cosine similarity, $q$ denotes \change{a given} sample, $k_+$ denotes positive pairs, $k_i$ denotes all negative pairs, and $T$ is a temperature hyperparameter that controls the alignment level of the learned embeddings~\cite{wu2018unsupervised}. \change{The positive pair has the same labels, while the negative pair has a different label. According to the analysis in~\cite{wang2020understanding}, a small $\sigma(q,k_+)$ represents alignment, and a large $\sigma(q,k_i)$ denotes uniformity. Alignment and uniformity are achieved by committing to the ``hardness''~\cite{tian2020makes,zhu2021improving} in positive and negative sampling, respectively. By ``hardness'' we mean hard samples that are similar and \change{difficult} to be distinguished from each other. We introduce positive and negative sampling in CL as follows: for positive samples, we generate or find similar samples from the given samples~\cite{shorten2019survey,feng2021survey,you2020graph}; for negative samples, we look for hard negative samples~\cite{schroff2015facenet,suh2019stochastic,robinson2020contrastive}. When an invalid sampling strategy is chosen, \change{the} beneficial hardness becomes harmful indistinguishability. An undesired local solution is satisfied when all \change{the} embeddings generated by the model are constant (i.e., model collapse). When we use the collapsed embeddings in fine-tuning, a negative impact \change{is observed,} and the performance of the model drops to the same level as the randomly initialized embeddings~\cite{zhu2021improving}. \change{Accordingly,} \change{selecting a }sampling strategy becomes \change{critical}. To our knowledge, only one work in the image domain~\cite{tian2020makes} \change{has proposed} a method for locating the best positive sampling strategy, and no general measure for promising sampling strategies \change{has been created}.}

\par \modify{We present an illustrative example of predicting house prices based on critical features including \textit{house style}, \textit{built year}, and \textit{street location}. We use the output embeddings of a house generated by a CL model as evidence to estimate its price. When houses share the same street location, built year, and style, they should have similar embeddings, indicating alignment. Conversely, when houses have any differences in these features, their embeddings should be distinguishable, indicating uniformity. To achieve both alignment and uniformity in CL, the concept of hardness is introduced. We showcase negative samples with hardness, which are houses sharing similar built years and street locations but possessing slightly different building styles and significantly different sale prices. We also showcase positive samples with hardness, which are houses also sharing similar features but possessing similar sale prices. Model collapse occurs when embeddings cannot accurately reflect house prices. For instance, when the embeddings of houses built in different years are identical, this leads to the loss of information about the year of construction and ultimately results in inaccurate price estimation.}

\begin{figure*}[h]
    \centering
     \vspace{-6mm}
    \includegraphics[width=\linewidth]{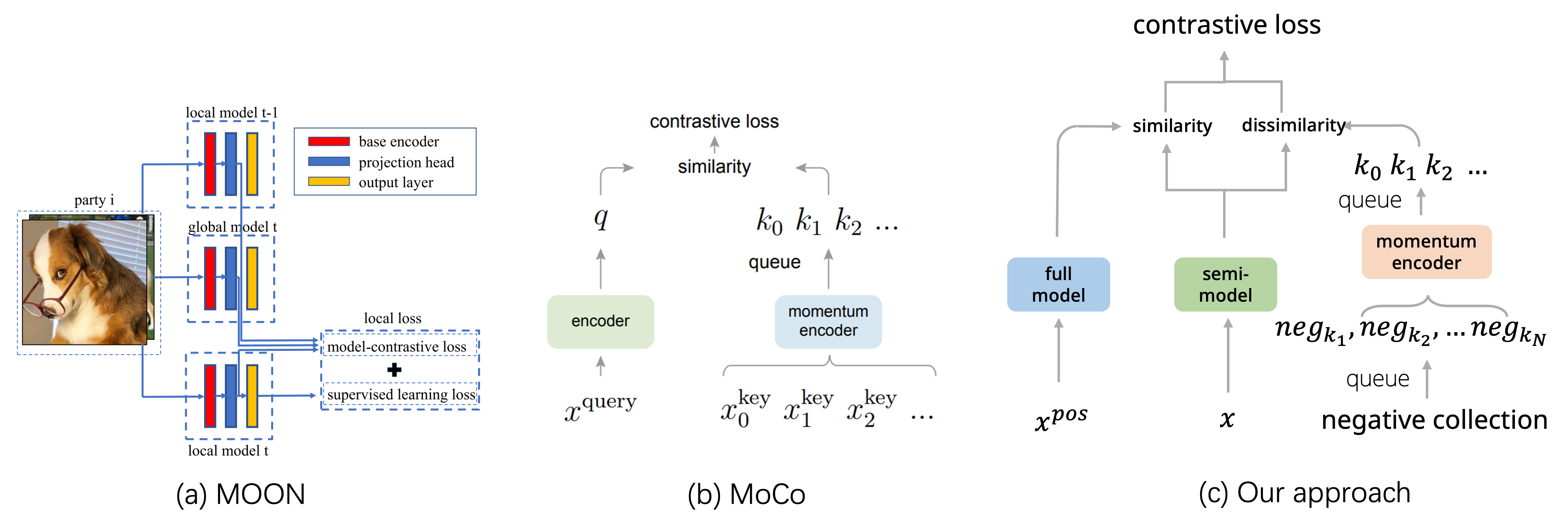}
    \vspace{-6mm}
    \caption{Model architecture: (a) MOON inspires our design in terms of positive sampling; (b) we adopt MoCo's negative sampling in our scenario; (c) our model, where $N$ is the batch size. The architecture of our model differs from MoCo in two aspects. First, our positive embedding is generated by another model. Second, the queue precedes the momentum encoder to construct a batch of negative samples. \modify{(a) is a reprint of \modifya{Figure 3} of \cite{li2021model} and (b) is a reprint of \modifya{Figure 1} of \cite{he2020momentum}.}}
    \label{fig:model}
        \vspace{-6mm}
\end{figure*}

\subsection{Architecture of Back-end Model}
\label{sec:backend}
\par \change{The back-end architecture is inspired by MOON~\cite{li2021model} and MoCo~\cite{he2020momentum}, for which we first introduce them.}

\par \change{\textbf{MOON.} The architecture of MOON is shown in~\autoref{fig:model}(a)), where a party $i$ implies a batch of data, the local and global models share the same architecture (dashed box) but are provided by different data, \change{and} $t$ represents the current epoch. The global and local models are simultaneously \change{trained} by \change{using} all data and partial data from a party. Thus, the global model is usually more accurate, and the output of both models constructs a positive pair, while the negative pair is generated by the local model in the last epoch ($t-1$). MOON optimizes the local model by minimizing the sum of the contrastive loss of the two pairs and the supervised loss depending on the downstream task.}

\par \change{\textbf{MoCo.} The architecture of MoCo is shown in~\autoref{fig:model}(b)), where $x^{query}$ is augmented from a given instance $x$, the encoder and the momentum encoder are with the same neural network with different update policies, a queue going through the entire dataset outputs $x^{key}$, and $q$ and $k$ are embeddings. The momentum encoder is updated by the small weights of the gradients from the encoder; $q$ and the embedding of $x$ construct a positive pair, while all $k$ and the embedding of $x$ construct a negative pair. MoCo optimizes the encoder by minimizing contrastive loss.}

\par \change{\textbf{Architecture of our model.} The way MOON constructs positive and negative pairs inspires us that a positive pair of CL could be embeddings of diverse models trained from different data. The momentum encoder in MoCo provides us with a prototype to generate embeddings of negative pairs that cover a variety of data and differ from the original embedding.} When we transfer \change{the} global and local models in MOON to complete and incomplete data and use the momentum encoder in MoCo to generate negative embeddings, the label prediction task in our case can be formulated as a subproblem of CL: \textit{For an incomplete sample, the ML model should learn the similarity of its complete counterpart and the dissimilarity between any other samples.} \autoref{fig:model}(c) shows the specific model architecture of our approach, \change{which includes \textbf{positive sampling} and learning similarity for alignment, \textbf{negative sampling} and learning dissimilarity for uniformity, and \textbf{contrastive loss} construction (i.e., optimization objective)}. Specifically, we first measure the sample distance and sample the hard positive pairs. Then, a new adaptive MoCo technique is used to identify \change{the} hard negative samples \change{in the case of incomplete data}. Finally, the similarity between positive pairs and the dissimilarity between negative pairs constitute the contrastive loss. Our objective function is a combination of task and \change{contrastive} losses \change{to further consider the downstream task.} The details will be explained in the following subsections.

\subsubsection{Positive Sampling}
\label{sec:alignment}
\par In our scenario, the positive pair $x^{pos}$ (\change{\autoref{fig:model}(c)}) is used as the ``complete form'' of the given data. The complete form of the data does not contain any missing values and dimensions (i.e., full data). \change{Meanwhile,} the given data come from the data after removing some feature dimensions (i.e., semi-data). We remove the unselected features (usually features with high missing rates) to generate semi-data, which \modify{come} from two types of raw data: 1) raw data without missing values or 2) raw data with missing values. In the former case, the positive pairs of the semi-data are naturally the original data. \change{In the latter case, for a semi-data, we score all the full data by similarity, \change{with} the highest \change{being} the positive sample.}

\par \change{\autoref{eq:dist_pos} measures the label and feature similarities between the semi-data and the full data.} We follow the definition of hard positive samples (i.e., high label similarity with small embedding distance)~\cite{xuan2020improved} in the classification and transfer to the regression (\autoref{eq:dist_pos}), where $\sigma$ denotes cosine similarity, $label_{s}$ and $label_{f}$ denote the labels of the semi-data and full data, \change{$x$ stands for 1) ``embedding'' representation: the embedding generated by the current model, or 2) ``raw'' representation: the feature vector of the raw data, $x_s$ and $x_f$ being $x$ for semi-data and full data. The former representation of $x$ is preferred when we expect samples with similar embeddings to be positive, and the latter one is preferred when we expect samples with similar features to be positive. In addition, the second representation of $x$ is particularly useful when the pre-trained model \change{poorly} performs.} We normalize the values of $abs(label_{s}-label_{f})$ and $abs(label_{a}-label)$ to values between $0$ and $1$.

\begin{equation}
    \begin{aligned}
    \label{eq:dist_pos}
    score_{p} = \sigma(x_{s}, x_{f}) - \frac{abs(label_{s}-label_{f})}{\max\{abs(label_{s}-label_{f})\}}
    \end{aligned}
\end{equation}

\par However, the samples found in the limited search space \change{may also be} infeasible positive pairs. For example, a semi-data is different from all the full data, and the closest full data can hardly be regarded as the complete form of the semi-data. \change{Accordingly,} we set a condition that when $score_{p}>\mathbb{E}_{(s,f)\sim P}[\sigma(x_s,x_f)]$, where $P$ denotes all positive pairs, no suitable positive sample is found, and we set the coefficient $\mu$ in \autoref{eq:loss} (introduced later) to $0$. By default, $\mu$ is $1$ when entering full data and $0.5$ when entering semi-data because its positive sample comes from another part of the data compared with the full data. \change{Specifically,} $\mu$ is used to adjust the contribution of the discriminant sample and stabilize the model training. Finally, the selected positive samples are fed into the full model to produce the embedding $k_+$ in \autoref{eq:con}.

\begin{figure*}[h]
    \centering
    \includegraphics[width=\linewidth]{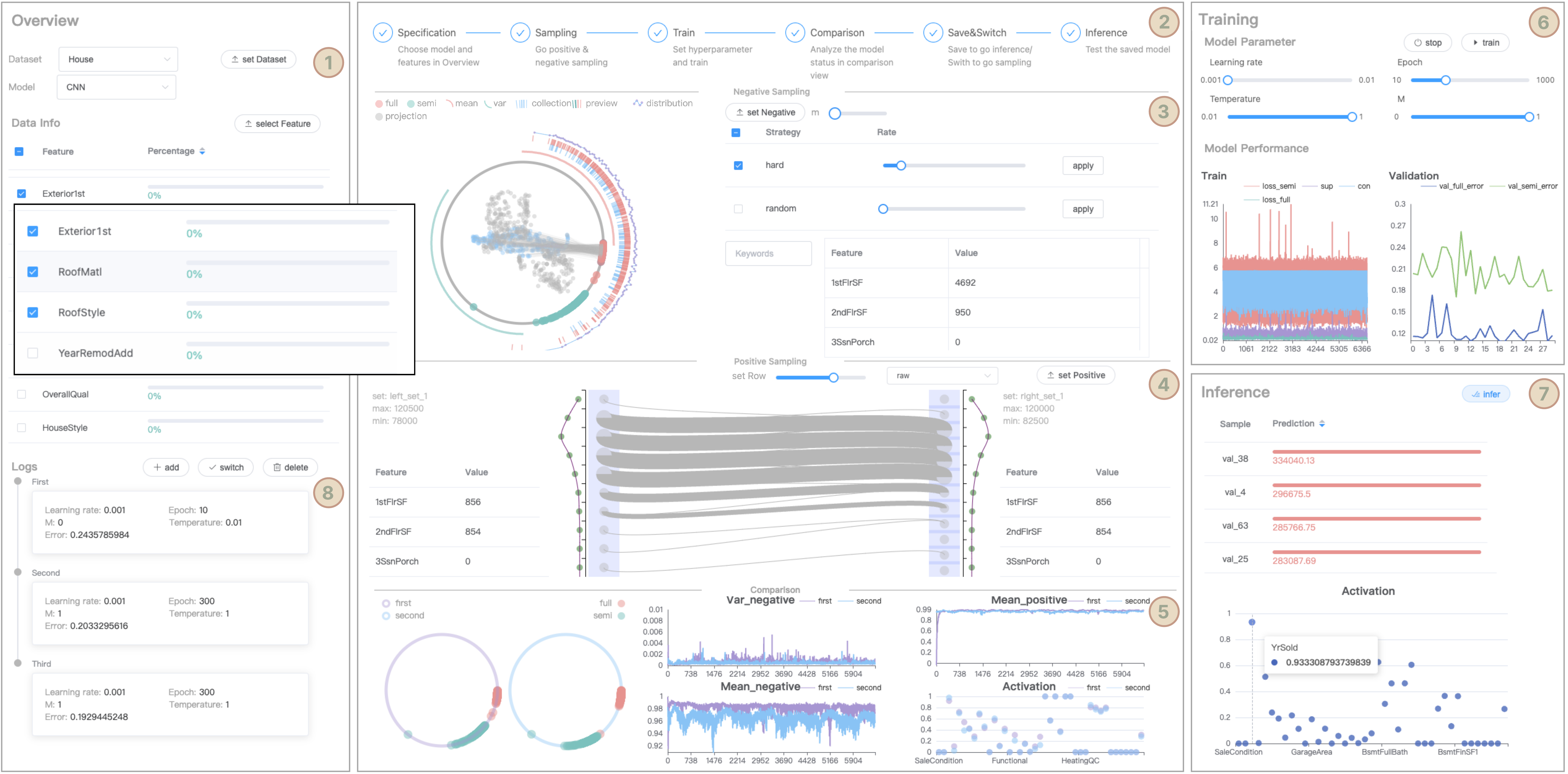}
    \vspace{-6mm}
    \caption{(1) The overview view enables the user to select various datasets and models, displays statistical information on the missing rate, and permits the specification of feature dimensions for the subsequent procedures. (2) The guidance view shows the current process and prompts for the next step. (3) The negative sampling view includes a circle-based glyph for examining the embedding distribution of the data records (right) and a control panel for further selecting the appropriate negative collection through the negative sampling strategies (left). (4) The positive sampling view displays the relationship between the data and its positive sample via a bipartite glyph and supports interactive adjustments. (5) The comparison view allows comparison and diagnosis during the training phase of both models. (6) The training view allows specifying hyperparameters and performing supervision during the training phase. (7) The inference view shows the predicted values of each test data and their activation. (8) Logs support model saving and switching between different trained models.}
    \label{fig:teaser}
        \vspace{-6mm}
\end{figure*}

\subsubsection{Negative Sampling}
\label{negative sampling}
\par The negative sample \change{for a given data point} is the data point with a different label. MoCo automatically includes the fed data as part of the negative samples \change{for each batch}. \change{The total set of negative samples is interactively given with the support of the front-end visualization described later, rather than being drawn from the batch data and potentially covering the entire dataset, to induce domain knowledge.} This modification helps the experts \change{in specifying} the partition of the data they want the model to get rid of.
\par \change{To facilitate the experts in constructing the total set,} we initialize two strategies for negative sampling, namely \textbf{random sampling}~\cite{wu2018unsupervised} and \textbf{hard negative sampling} (i.e., low label similarity with small embedding distance~\cite{xuan2020improved}) and transferred to the regression in \autoref{eq:dist_neg}. The notation of \autoref{eq:dist_neg} is similar to that of \autoref{eq:dist_pos}, \change{except that suffix $a$ represents the random anchor points for other samples.} The total set of negative samples can be interactively adjusted, for example, by adding a new negative sample or removing one after the initial sampling.

\begin{equation}
    \begin{aligned}
    \label{eq:dist_neg}
    score_{n} = \sigma(x_{a}, x) + \frac{abs(label_{a}-label)}{\max\{abs(label_{a}-label)\}}
    \end{aligned}
\end{equation}

\par \change{Thereafter,} we generate the embedding of negative samples according to MoCo, which consists of two parts (\autoref{fig:model}(c)). First, MoCo maintains a queue to extend the sources of negative samples. In our model, the semi-data and full data have different dimensions, and our semi- and full-queue follow the same update mechanism. In each training batch, the current sample in the queue is regarded as a negative sample. The queue is then updated by the selected data, first in, first out. Subsequently, the negative samples are fed to the momentum encoder to generate the negative embedding $k$ for~\autoref{eq:con}. We also have two momentum encoders in our model. After the embedding is generated, the momentum encoder is updated by the parameters of the full/semi model only, instead of backpropagation:
\begin{equation}
    \hat{\theta}_k = \hat{\theta}_k+(1-m)\theta_k, k=\{s,f\},
\end{equation}
where $\theta_s$ and $\theta_f$ denote the parameters of the semi and full model, respectively; and the $\hat{\theta}_s$ and $\hat{\theta}_f$ denote the parameters of the semi and full momentum encoder, respectively. When $m=1$, the semi- and full momentum encoders are equal to the semi- and full model, respectively. In summary, the hardness in the negative sampling \change{can be adjusted in three ways}: 1) the sampling strategy; 2) the samples to which the strategy applies; and 3) the hyperparameter $m$ in the momentum encoder.

\subsubsection{Objective Function}
\par In the training phase, the loss function of the semi-model includes \textit{task loss} ($L_{task}$) and \textit{contrastive loss} ($L_{con}$):
\begin{equation}
    L=L_{task}(q,y|\theta_f)+M\cdot \mu \cdot L_{con}(q|\theta_s).
    \label{eq:loss}
\end{equation}
The task loss depends on the downstream task (e.g., the MSE loss of the house price prediction \change{scales from zero to one}). The observed data $q$ and its label $y$ are fed into the task loss to optimize $\theta_f$, i.e., the full model. Contrastive loss (\autoref{eq:con}) uses the resulting positive and negative embeddings (depending on the input $q$) to improve the performance of the semi-model $\theta_s$. $M$ is the weight of $L_{con}$ and $\mu$ is the coefficient for the positive pairs mentioned in \autoref{sec:alignment}. Then, the full model is trained together with the semi-model. In the inference phase, the separately trained semi-model provides the prediction results for the downstream task.

\section{Front-end Visualization}
\par Based on the preceding requirements and the proposed CL-based framework, we develop \textit{CIVis}, a visual analytics system (\autoref{fig:teaser}), which visualizes a set of information to facilitate the interactive selection of positive and negative sampling strategies and handle \change{the issue of modeling observed data with missing values}. \change{We went through several iterations during the development process. The initial version of the system was designed based on the team's requirements and contained various functionalities to support flexible operation during training. Later, the question of whether the interface needed to be simplified was raised, followed by a lengthy discussion about reducing unnecessary elements and adding user guidance. The discussion resulted in the development of two design principles for the interface: 1) visual updates after the operations should be automatic to reduce user manipulations and direct their attention to the next step, and 2) visual design of a progress bar is necessary to provide guidance for users and direct them to different stages of model development. Therefore, we applied the two principles in the final version of the visual analytics system.}
\par The whole pipeline can be summarized in the following steps: \textit{data preparation}, \textit{model configuration}, \textit{training, and inference}. The layout of \textit{CIVis} allows the user to iterate through the entire pipeline from left to right. The leftmost column is like a control panel (\autoref{fig:teaser}(1) and (8)). The user can select the dataset and the appropriate model structure, understand the overall missing features and select the necessary ones (\textbf{R.1}), recover the trained models, and switch between them. The middle column (\autoref{fig:teaser}(2), (3) and (4)) is the CL-oriented area in which we design glyphs to reveal the sampling and learning procedures for CL (\textbf{R.2}). In the comparison view (\autoref{fig:teaser}(5)), the multi-aspect metrics plot the training procedure (\textbf{R.3}). The rightmost column (\autoref{fig:teaser}(6) and (7)) refers to the training and inference phases. The user can adjust hyperparameters, such as learning rate, monitor the training process in the training view, and check the predictions for each unlabeled test data in the inference view for increasing confidence (\textbf{R.4}). In the following subsections, we introduce the visual design of the negative sampling view, the positive sampling view, the training and comparison view, the inference view, and the guidance view.

\subsection{Negative Sampling View}
\label{negative sampling view}
\par
\begin{figure}[h]
    \centering
    \includegraphics[width=\linewidth]{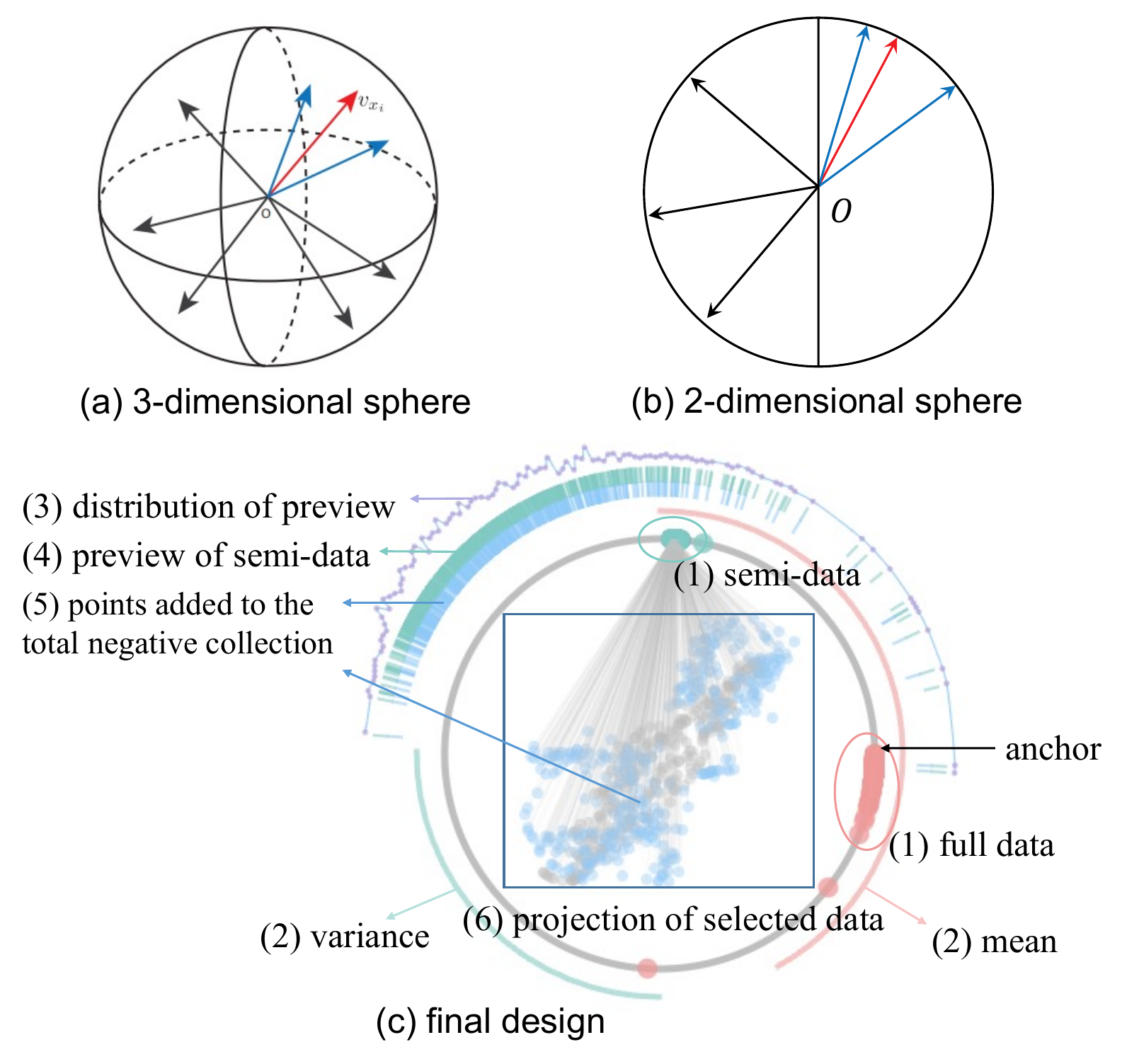}
    \vspace{-6mm}
    \caption{Design alternatives (a-b). (c) Visualization of the circle-based negative sampling view: From outermost to innermost, (1) indicates semi-data and full data; (2) indicates the variance and mean of the angle of the selected points; (3) indicates the distribution of the preview; (4) is an enlarged preview of the selected green/red points; (5) indicates the points added to the total negative collection; and (6) shows the projection of the selected data, where the points added to the negative sample collection are in blue. \modify{(a) is a figure reprinted from \modifya{Figure 2} of \cite{wu2018unsupervised}. (b) is the counterpart of (a) in the 2-dimensional space.}}
    \label{fig:neg}
        \vspace{-6mm}
\end{figure}

 The negative sampling view (\autoref{fig:teaser}(3)) yields the total negative collection used for CL training. Previous studies~\cite{liu2017sphereface,wu2018unsupervised,zhu2021improving} demonstrated their validity by projecting high-dimensional latent representations onto a sphere (\autoref{fig:neg}(a)). Inspired by their designs, we propose a circle-based visualization method to facilitate experts to inspect the embedding distribution of data records and further select the appropriate negative collection (\autoref{fig:neg}(c)) (\textbf{R.2}). We also show the raw information of the selected data (i.e., the values of the feature dimensions in the bottom right of \autoref{fig:teaser}(3)) \change{to induce domain knowledge.}

\par \textbf{Visualization.} In \autoref{fig:neg}(c), we randomly choose one embedding of the data record as the ``anchor'', drawn on the rightmost point with degree $0$. The other embeddings are drawn on the innermost circle, where the angular difference from the anchor is the arccosine of the cosine distance. The green and red colors indicate semi-data and full data embeddings, respectively (\autoref{fig:neg}(1)). The role of visualization is to perceive uniformity, so the absolute position of points is meaningless. According to the uniformity principle, embeddings close to each other should receive more attention. To help check for ``uniformity'', we append two arcs with the arc length indicating the variance on the left and mean on the right (\autoref{fig:neg}(2)) to imply uniformity of the selected data records. These two arcs are from the bottom to the top and from the top to the bottom. However, potential visual clutter may appear on the innermost arc (i.e., many points are crowded together) \change{when the model is not robust enough to discriminate the data records well}. In this case, the nodes in the innermost circle will overlap, and their distribution of angular scales is difficult to examine in a small area. \change{To solve this problem, we plot the selected points on a larger scale as a preview (\autoref{fig:neg}(4)) and draw a purple curve (\autoref{fig:neg}(3)) to visualize the density distribution of the selected points along the preview.} As mentioned in \autoref{negative sampling}, we provide two sampling strategies: random sampling and hard negative sampling. The sampled points are added to the total negative collection, indicated by the blue arcs around the inner part of the preview (\autoref{fig:neg}(5)). In addition, we project the original data records onto a 2-dimensional plane by \textit{t-SNE}~\cite{van2008visualizing} \change{inside the innermost circle} (\autoref{fig:neg}(6)). Upon user interaction with the innermost circle, the blue projected node is linked to its corresponding point on the innermost circle, and the blue projected nodes are also included in the negative sample set.

\par \textbf{Interaction.} The negative sampling view presents the nodes and projections within the innermost circle. The selection of points of interest (PoI) allows the negative sampling view to show information such as the corresponding link to the projection, mean, and variance of the arcs, and a preview on the outermost circle. Only the projection nodes that correspond to the PoI are visible. The expert can select a strategy and adjust the sampling rate to negatively sample the PoI, resulting in the sampled nodes turning blue in the projection. Furthermore, the raw data corresponding to any point in the projection area can be displayed in the table on the right by clicking on it (\autoref{fig:teaser}(3)). The expert can also search for a specific feature value by entering a keyword. Finally, the momentum coefficient in MoCo, which determines the update speed of the momentum encoder, can be adjusted using a parameter slider bar named ``m''. The expert can configure the negative sample collection by clicking the \textit{set Negative} button.


\par \textbf{Design Alternative}. The design is derived from the sphere~\cite{liu2017sphereface,wu2018unsupervised,zhu2021improving} commonly used in the embedding analysis. \autoref{fig:neg}(a) depicts a 3-dimensional sphere, but a 128D unit sphere is implied in its original paper~\cite{wu2018unsupervised}. The purpose of the ``glyph design'' is to facilitate understanding of their method. \change{However, this 3-dimensional design does not facilitate direct access to information, such as distances compared with common 2-dimensional visualizations.} Accordingly, we improved the design of \autoref{fig:neg}(a) to \autoref{fig:neg}(b), where an arrow indicates the embedding projected on a 2-dimensional unit circle. The graphics directly shows the cosine distance between embeddings. Although an arrow naturally represents a vector, a node on the circle is sufficient to represent an embedding. In addition, we need to present the projection of the original data, as shown in our final design (\autoref{fig:neg}(c)).


\subsection{Positive Sampling View}
\label{positive sampling view}
\par The positive sampling view (\autoref{fig:teaser}(4)) is designed to generate and examine the positive samplings for each semi-data (\textbf{R.2}). Specifically, positive sampling is a many-to-many mapping, in which the semi-data look for the most similar full data, which is naturally a bipartite graph. However, directly drawing a bipartite graph with many nodes or edges often causes severe visual clutter, which will prevent experts from clearly understanding the mapping and further adjusting it according to their domain knowledge. Therefore, we develop a \change{Sankey-like visualization~\cite{riehmann2005interactive}} to describe many-to-many mapping relationships.

\begin{figure}[h]
    \centering
    \includegraphics[width=0.8\linewidth]{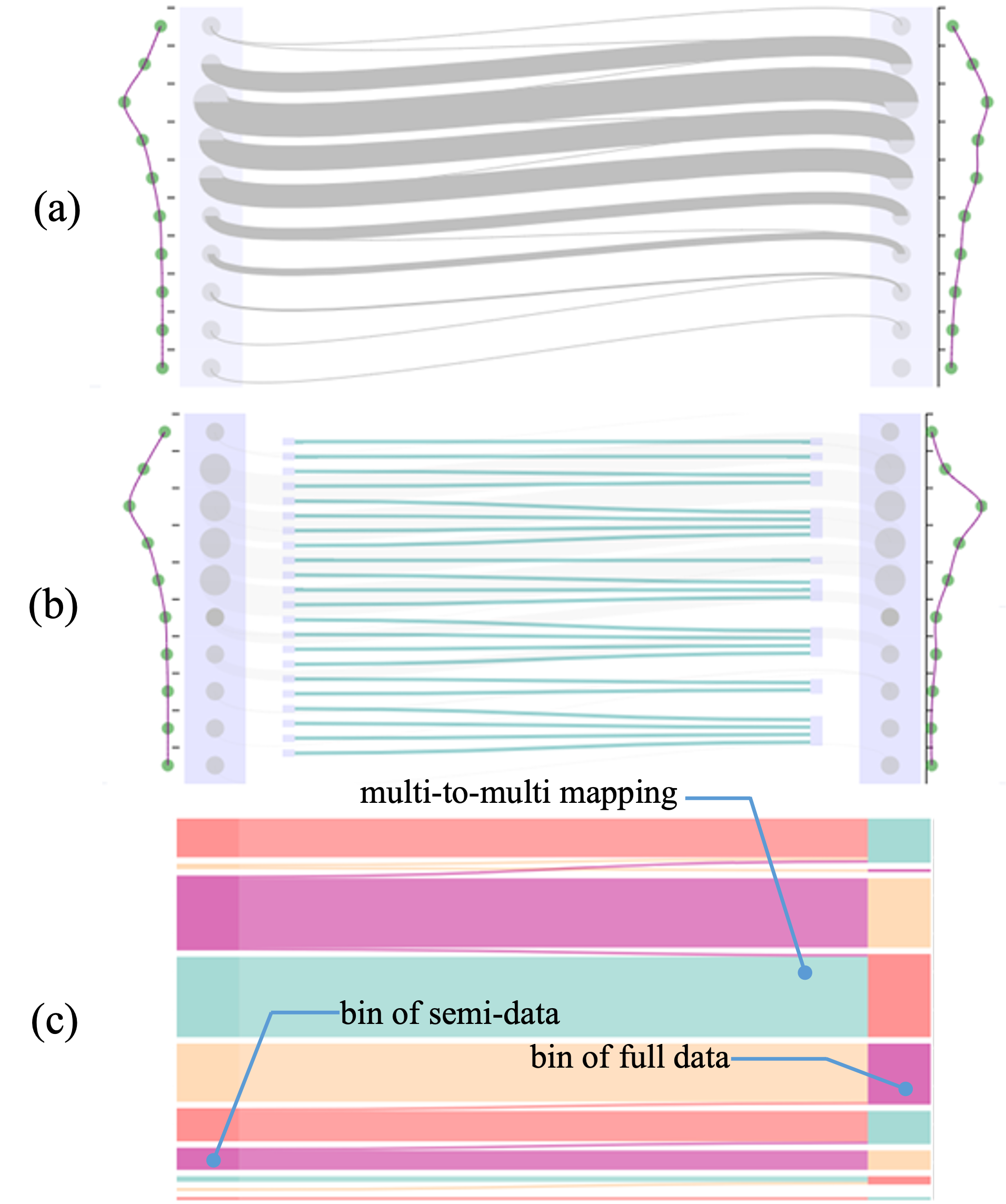}
    \caption{(a) The node on the left denotes a bin containing semi-data, and the node on the right represents a bin containing full data. The links on both sides represent a bin-to-bin mapping relationship. The node diameter and link width represent the number of mapping relationships. \change{(b) and (c) are design alternatives.}}
    \label{fig:pos}
    \vspace{-6mm}
\end{figure}

\par \textbf{Visualization}. To begin, we partition the semi-data and full data into $K$-labeled bins, where the node labels correspond to the bins, and maintain a bipartite graph. The selection of $K$ is determined based on experimental results for a specific dataset, which enables us to visualize the Sankey-based extended glyph clearly, as shown in \autoref{fig:pos}(b). In \autoref{fig:pos}(a), the node on the left side denotes a bin of semi-data. The diameter of the node is equal to the width of the corresponding intermediate link, which represents the mapping between bins. The width of the link indicates the amount of data in the bin that satisfies the mapping. We employ a Bézier curve~\cite{wells1990longman} to illustrate the links clearly, where the control points are located on the upper node on the left and the lower node on the right. A curve is drawn next to the nodes to display the amount of data in the bin, i.e., a histogram of the data labels. By clicking on the nodes, we can present the maximum and minimum values and feature averages in the bins.


\par \textbf{Interaction.} \change{The positive sampling view initializes the mapping obtained from the search method mentioned in~\autoref{sec:backend}. Two interactions are related to the generation of the default mappings. The \textit{set Row} bar is \change{used} to adjust the number of bins (i.e., the number of nodes on both sides) according to the downstream task (e.g., for classification), and the number of bins and the number of classes should be equal. Furthermore, the drop-down box provides \textit{embedding} and \textit{raw} options (introduced in~\autoref{sec:alignment}) to specify the representation used in the algorithm. Clicking the \textit{set Positive} button will finalize the mapping between the semi-data and the full data.}

\par \textbf{Design Alternative}. \change{We performed a functional simplification in the positive sampling view during the design iterations. The Sankey-like alternative (\autoref{fig:pos}(b)) describes a one-to-one mapping of an edge in the positive sampling view. Experts can adjust a mapping when they find an anomaly. However, adjusting a mapping requires at least seven steps, and checking the mapping one by one \change{is tedious and confusing for experts}. After discussion with the experts, we decided that the automatic method for the positive sampling (\autoref{sec:alignment}) provides a promising mapping for the positive sampling, so fine-grained adjustment is a dispensable requirement, especially when it requires an unacceptable manual effort. \change{Consequently,} we remove this option to reduce the number of operational steps in our system. In addition, we went through one design iteration when we used Sankey glyphs to draw the many-to-many bipartite graph in the positive sampling view (\autoref{fig:pos}(c)). The drawback of Sankey graphs is that the thin links \modify{and rectangles} are almost invisible. The narrow links typically lead to a bin with a range of different labels and reveal potential anomalies. Therefore, we improved the design by limiting the diameter and width of the nodes and links. Those thin but important links are clearly displayed, and the smaller nodes can be easily interacted with.}

\subsection{Training and Comparison View}
\label{training comparison view}
\par The training and comparison view (\autoref{fig:teaser}(5) and (6)) controls and supervises the training phase (\textbf{R.3}). The expert can adjust hyperparameters, including \textit{Learning rate}, \textit{Epoch}, \textit{Temperature}, and $M$ in~\autoref{eq:con}, and control the training state via the \textit{train} and \textit{stop} buttons. Then, \textit{CIVis} updates the training-related metrics in real time.
\par In the training view, the training loss in \textit{Train} and the validation error in \textit{Validation} are updated; in the comparison view, \textit{Var\_neg} indicates the variance of the negative scores (the cosine distance between the fed data and the negative collection), \textit{Mean\_neg} indicates the mean of negative scores, and \textit{Mean\_pos} denotes the mean of the positive scores. \change{After the training is completed, the circle glyph on the left side of the view and \textit{Activation} in the lower right corner will be updated (\autoref{fig:teaser}(5)).} The encoding of the circle glyph is the same as the innermost circle of the circle-based glyph in \textit{negative sampling view}. The \textit{Activation} describes the contribution of a feature to the predictive result.

\par \change{The current and next rounds of the training yield the ``first'' and ``second'' models. The above-mentioned visual cues in the comparison view are indicated in purple and blue to facilitate comparison. Hovering any figure in the training and comparison view puts the curve in the front to avoid visual clutter, and the exact value is displayed by the tooltip.} At this point, the trained model (i.e., semi-model) has gone through the entire pipeline of \textit{CIVis} and can be saved by clicking on the \textit{add} button, which appends a single line to \textit{Logs} (\autoref{fig:teaser}(8)). Clicking on a single line switches the system to the corresponding model. Meanwhile, clicking on the \textit{delete} button also deletes the single line of records. Thus, \textit{CIVis} allows experts to analyze and refine a model in an \change{iterative and} interactive way.

\par \textbf{Design Alternative}. In the comparison view, the layout of two circular glyphs undergoes one design iteration when they share the same center but have different radii. This alternative design shows advantages in comparing the distribution of points within the same central angle. However, it does not take into account the visual misinterpretation caused by the different radii, for which the inner circumference covering the same \change{angle} is naturally shorter. In other words, even if the inner distribution of points is similar to the outer one, this alternative can cause the misconception that the inner distribution is visually narrower. Therefore, we improved the layout to ensure that the circle glyphs had the same shape and placed them side by side, which eliminated the visual misunderstanding and caused little inconvenience when comparing point distributions within the same range.

\subsection{Inference View}
\label{inference view}
\par \textit{CIVis} preserves a trained semi-model that experts can use in the inference phase (\autoref{fig:teaser}(7)) (\textbf{R.4}). Although in reality, we cannot guarantee which features are missing \change{in the inference phase}, and no ground truth exists, we discard those data records that have missing values in the selected features for experimental simulation. The expert can re-run the feature selection process if the discarded data unexpectedly appear. The remaining data records (i.e., those with values in the selected features) are fed into the trained semi-model, and the \textit{inference view} will display the predicted values and activation values \change{based on GradCAM~\cite{selvaraju2017grad}}. \change{In the upper part, each row refers to the prediction of a data record; in the lower part, each point refers to a feature importance.} The expert can further check whether the appropriate features are activated to verify the predicted values.

\subsection{Guidance View}
\par \textit{CIVis} includes the entire process of training an ML model and many components must be integrated. \modify{In order to enhance clarity and mitigate confusion in the utilization of the system, prescribing guidance is integrated into our approach, as recommended by Ceneda et al.~\cite{ceneda2019review}. This prescribing guidance suggests view-by-view actions for optimal utilization of the system.} The completion of one view automatically triggers the update of the next view, and the guidance view at the top of the positive sampling view (\autoref{fig:teaser}(2)) shows the current progress. For example, when the selection of models, datasets, and features change the positive and negative sampling views, the guidance progress goes from \textit{Specification} to \textit{Sampling}. In particular, clicking \textit{switch} in \textit{Logs} triggers an update of the positive and negative sampling views, and the guidance progress returns to the \textit{Sampling} step because \textit{CIVis} supports progressive development. Finally, when experts test the saved model in the inference view, they reach the end of the pipeline. \modify{Across the entire pipeline, the updates that are triggered ought to attract the users' attention and direct them towards the subsequent view that requires their focus.}

\begin{figure*}[h]
    \centering
    \includegraphics[width=\linewidth]{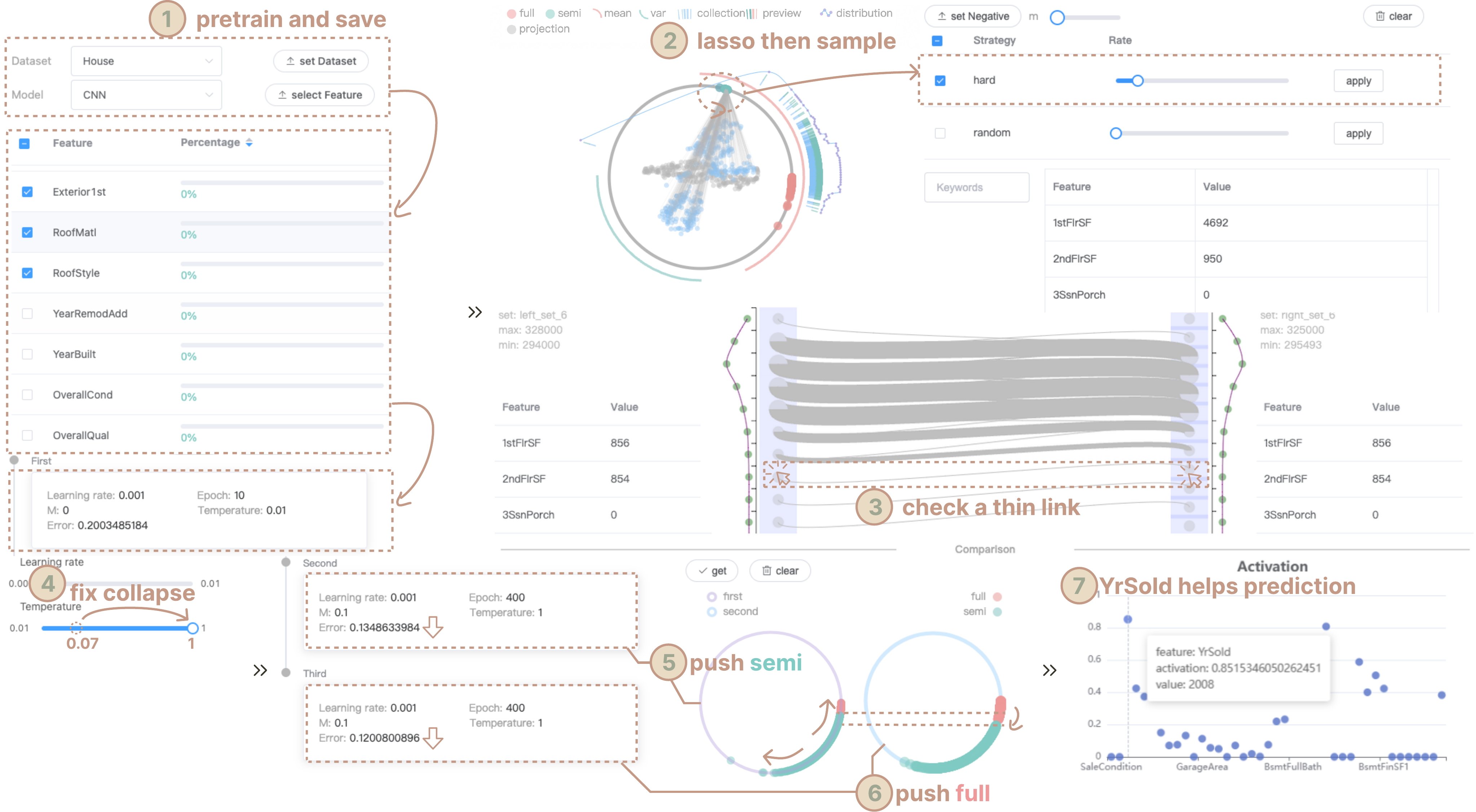}
    \vspace{-6mm}
    \caption{Interaction and observation in \change{usage scenario I}. The notation $1-3$ demonstrate how to configure positive and negative sampling based on visual cues; notation $4-7$ highlight and explain the benefits that \textit{CIVis} brings. The loss curves in (4) look like an area because the loss sharply and frequently jumps up and down.}
    \label{fig:case1}
        \vspace{-6mm}
\end{figure*}

\section{Evaluation}
\par In this section, we evaluate the effectiveness of \textit{CIVis} from several aspects of the visualization community~\cite{isenberg2013systematic}. First, we describe two usage scenarios with our collaborative experts (E1-E3) who participated in our user-centered design process. Second, we conduct a quantitative experiment comparing \textit{CIVis} with several selected data imputation baseline methods. Third, we interview the experts to obtain their feedback on \textit{CIVis}. \change{Finally, we conduct a qualitative user study to further evaluate the effectiveness of \textit{CIVis}.}

\subsection{Usage Scenario I: House Price Prediction}
\par The modeling data of the real estate industry mentioned in the observational study have confidentially rules and cannot be used as experimental data. As an alternative, E1 suggested that we could take a publicly available house price dataset~\cite{HousePrice}, which is similar to their scenario. The dataset includes a feature dimension of $79$ and the task is to predict house prices using a convolutional neural network (CNN). In particular, this usage scenario illustrates how the expert (i.e., E2) understood the training phase, steered the CL-based framework, and induced his knowledge into the trained model to deal with the issue of modeling observed data with missing values with \textit{CIVis}. The value types of features include numbers and strings. Consequently, we apply dummy encoding in \textit{Pandas}~\cite{mckinney2011pandas}. Specifically, if a feature dimension has four possible values, then we extend the feature to four dimensions and use a one-hot encoding. We split the first $10\%$ of the dataset as the validation set. We also split a test set ($10\%$) for the inference phase \change{to facilitate the evaluation of the model performance}. The CNN model consists of three 2-dimensional convolutional layers and three fully connected layers.

\par E2 first specified the features to be used and the ``to-be-improved'' model, as shown in~\autoref{fig:case1}(1). After loading the dataset and CNN model into \textit{CIVis}, E2 checked the overall missing data rates in \change{\textit{Overview}}. To avoid missing data, he eliminated $15$ features with missing rates greater than $0$ and further eliminated $19$ \change{features that were not clearly defined, such as ``MiscVal'' representing miscellaneous feature values.} Finally, the \change{$45$} features were retained. Then he clicked the \textit{select Feature} button and waited for pre-training in \textit{CIVis}. After pre-training, he clicked the \textit{add} button in \textit{Logs} to save the pre-trained semi-model as a baseline. Note that \textit{CIVis} pre-trains the full model by all features and the semi-model by selected features.

\par \textbf{Negative Sampling Interaction.} When the guidance view proceeds to \textit{Sampling}, the negative sampling and positive sampling views are automatically updated, and the expert moved to the negative sampling view. \change{Initially, the green dots (the dashed circle in \autoref{fig:case1}(2)) cover a particularly narrow range, which means that the variance of the semi-data is small. Then, E2 lassoed all the green dots \change{for verification}. The updated variance (lower left corner of \autoref{fig:case1}(2)) is relatively short, and the updated preview (right side of \autoref{fig:case1}(2)) draws a green region where the semi-data are dense.} The expert then concluded that the current embedding distribution violates the principles of ``alignment'' and ``uniformity''. \change{He then wanted to refine the distribution \change{to ensure that} at least one significant gap \change{exists} between points (e.g., the red points in \autoref{fig:teaser}(3)) and added the samples to the negative collection, ``\textit{samples added to the negative collection should be pushed away on the circle and have a larger variance}''.} Accordingly, he again lassoed all the semi-data, ticked the ``hard'' sampling strategy, dragged the \textit{Rate} slider to $0.6$, and clicked the \textit{apply} button, which automatically added $60$\% of the selected samples into the negative collection (\autoref{fig:case1}(2)). \change{The updated blue arc in the preview (on the right side of \autoref{fig:case1}(2)) presents the added samples. Their} density indicates that the distribution of the added samples is reasonable (i.e., the dense areas in the preview are also dense in the corresponding areas of the blue arcs). Therefore, he clicked the \textit{set Negative} button and moved on to the positive sampling view.

\par \textbf{Positive Sampling Interaction.} \change{E2 then went to the positive sampling view.} He witnessed some thick links in the middle of the view \change{(\autoref{fig:case1}(3))}. Then, he clicked the corresponding bins of the thick links to check the label ranges. \change{Since their feature averages (left and right columns of \autoref{fig:case1}(3)) and label ranges (upper left and upper right corners of \autoref{fig:case1}(3)) are similar between the semi-data and positive bins}, \change{E2 was satisfied with the default results. So he just clicked on the \textit{set Positive} button and the guidance view proceeded to \textit{Train}.}

\begin{figure}[h]
 \centering 
 \includegraphics[width=\linewidth]{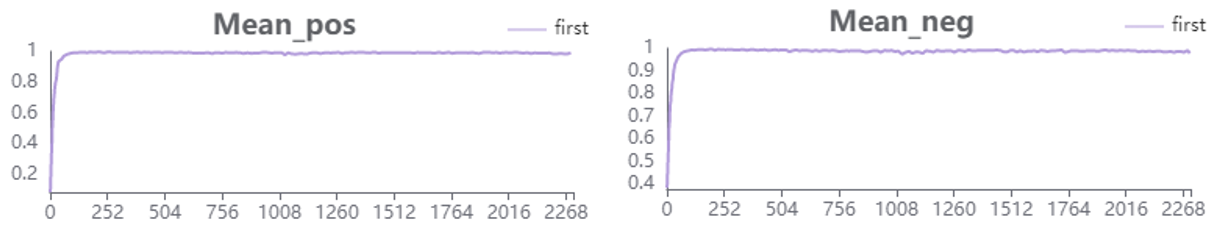}
      \vspace{-6mm}
 \caption{Model collapse. The mean of cosine distance is one.}
 \label{fig:collapse}
   \vspace{-3mm}
 \end{figure}

 \begin{figure*}[h]
    \centering
    \includegraphics[width=\linewidth]{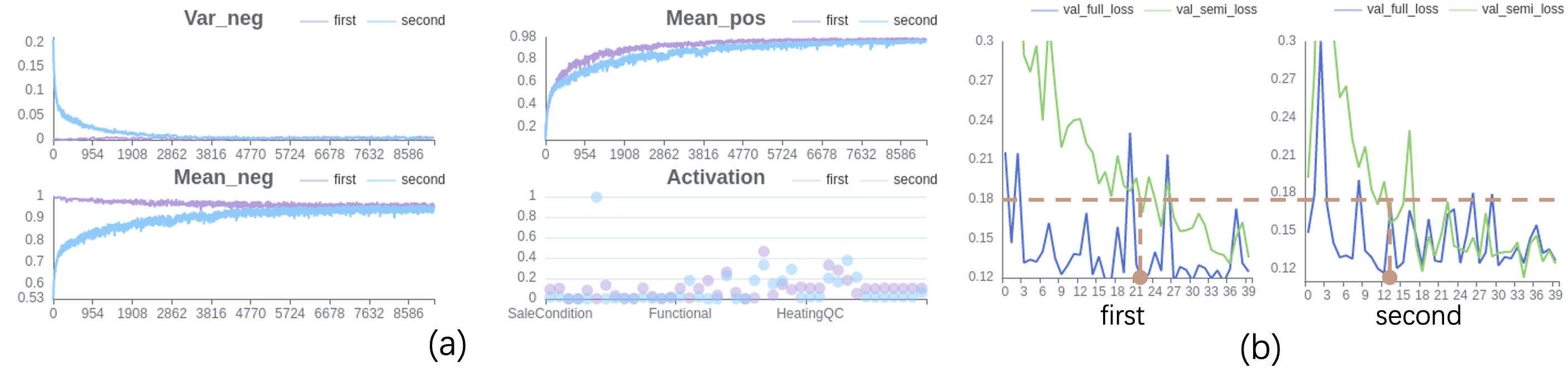}
        \vspace{-8mm}
    \caption{\change{Explanation of the training phase of CL. The difference between the first and the second rounds are with only the negative collection. (a) reveals the distance change between positive and negative pairs during training and \textit{Activation} describes the feature importance. (b) plots the loss curves of the full and semi-model during training.}}
    \label{fig:num}
        \vspace{-6mm}
\end{figure*}
 
\par \textbf{Training and Model Collapse.} After specifying the negative collection and checking the many-to-many mapping to find the positive samples, E2 dragged the slider in the training view to set the hyperparameter based on his domain experience. The \textit{Temperature} parameter is designed to amplify the cosine distance between a sample and its positive and negative samples. \change{The expert set the \textit{Temperature} parameter to $0.07$ according to MoCo and set a small number of epochs of $100$ to see its effect on the model training.} The pre-trained full model would be trained at the same time. After clicking the \textit{train} button, \textit{Var\_neg}, \textit{Mean\_neg}, \textit{Mean\_pos} in the comparison view, and \textit{Loss}, and \textit{MSE} in the training view are updated in real time. With a few epochs, \textit{Var\_neg} goes down to around zero. \textit{Mean\_pos} and \textit{Mean\_neg} quickly rise to one, implying that the training embeddings are similar (\autoref{fig:collapse}). The expert concluded that model collapse occurs. He then proposed a hypothesis that a smaller \textit{Temperature} might break the training phase. To validate his hypothesis, he set \textit{Temperature} to one and repeated the above-mentioned procedure (\autoref{fig:case1}(4)). After clicking the \textit{train} button, the expert found that \textit{Mean\_neg} and \textit{Mean\_pos} started to slowly evolve, \change{just like the first few epochs in \autoref{fig:num}(a)}; thus, the problem of the model collapse was solved.

\par \textbf{Training with the Semi-data.} \change{After fixing the model collapse, E2 started another round of training, adding more epochs, to see how the model converged in the updated distribution and activation in the comparison view. The purple curve in \autoref{fig:num}(a) depicts the training phase of the CL. The \textit{Mean\_neg} ranges from $1$ to about $0.95$, implying that negative pairs are \change{pushed away} with the contribution of CL. The \textit{Mean\_pos} rises from close to $0$ to $0.98$, implying that the distance between the positive pairs becomes smaller with respect to another CL contribution. In summary, the purple curve reflects the training phase exactly as the CL aims to push away the negative pairs and pull close the positive pairs.} \change{The distribution of the semi-data is wider compared with} the distribution produced by the pre-trained model (green in \autoref{fig:case1}(2)), which is consistent with the expert's purpose of pushing them apart (purple circle in \autoref{fig:case1}(5)). In the training view, the expert observed that the contrastive loss slowly decreases, and the validation MSE of the semi-model catches up with the MSE of the full model (\autoref{fig:num}(b)). Clicking the \textit{add} button in the \textit{Logs} records the hyperparameter for this training and the MSE error ($0.13486$).

\par \change{\textbf{Training with the Full Data.} The above training adds only the semi-data to the negative collection (the first run). E2 wanted to compare the case \change{wherein} the full data were added to the negative collection (the second run). \change{He clicked on the first row in \textit{Logs} and then clicked on the \textit{switch} button to restore the pre-trained model as a baseline, \change{similar to} the first run. The guidance view returns to \textit{Sampling}, and the positive and negative sampling are automatically updated.} Nevertheless, the negative collection consisting of only the full data increased the possibility that an instance of the full data is the positive pair while within the negative collection, \change{resulting in} an abrupt increase in CL loss and instability of the learning process. Therefore, he used the same procedure as above and lassoed the full data at a sampling rate of $90\%$, adding a sample of $470$ to the negative collection. After another round of training, he clicked the \textit{add} button in \textit{Logs} and found that the MSE \change{decreased} (i.e., from $0.13486$ to $0.12008$ as shown in~\autoref{fig:case1}(6)).}

\par Then, E2 moved to the comparison view (\autoref{fig:num}(a)) to find the reason for the decrease. When the blue curve in \textit{Mean\_neg} tends to be smooth, the blue curve in \textit{Var\_neg} tends to $0$, and \textit{Var\_neg} indicates the convergence degree. In \textit{Mean\_neg}, the blue curve starts from $0$, opposite to the purple curve, and reaches below the purple curve. According to E2, ``\textit{this observation yields three following insights''}: 1) The $0$ starting point of the blue curve indicates that embeddings of full data are initially distinguishable. Meanwhile, the $1$ starting point of the purple curve indicates that embeddings of semi-data are difficult to distinguish because they have fewer features compared with the full data. 2) In \textit{Mean\_neg} and \textit{Mean\_pos}, the purple curve below the blue curve indicates that all instances are more significantly different from the full data. \change{Accordingly, the loss of the semi-model decreases faster and finally reaches a lower level in the second round ((\autoref{fig:num}(b))).} Zhu et al.~\cite{zhu2021improving} also confirmed that discriminative positive samples can improve CL performance. 3) The blue curve up in \textit{Mean\_neg} means that the instances are getting closer to the full data in the negative collection, which is against the purpose of CL at first glance. However, the full data are in the negative collection, which is also a positive sample, so the training phase of CL finds a reasonable distance between instances and the full data. ``\textit{Through this scenario, I learned profoundly that CL is about trade-offs between alignment and uniformity, not simply pushing away the negative pair and pulling close the positive pair,}'' concluded E2. Meanwhile, the circle glyph in the comparison view (\autoref{fig:case1}(6)) shows a wider distribution of the full data compared with the first run. The blue and purple curves rise in \textit{Mean\_pos}. These observations lead to the same conclusion as the first run: the embedding of the data added to the negative collection becomes more discriminative, while the embedding of the positive samples becomes similar.

\par We summarize the contributions of \textit{CIVis}: 1) \textit{CIVis} helps the expert in understanding the training phase and deepen his understanding of CL; 2) \textit{CIVis} supports a highly interactive mechanism to generate discriminative samples and control their degree of discriminability (depending on how many negative samples are added); 3) the generated discriminative samples improve the semi-data performance.

\par \textbf{Inference.} After the training, E2 was more confident about the trained model. Then, he clicked the \textit{infer} button to trigger inference. Subsequently, he proceeded to verify the predictions in the inference view. Specifically, he clicked on the row with the smallest house price prediction, and the highest activation was \textit{YrSold} (the year the house was sold) (\autoref{fig:case1}(7)). ``\textit{In my opinion, the year the house was sold should be similar to the year the house was built, and older houses may have problems such as safety reasons and outdated facilities that may affect the price of the house}'', said E2. He also observed several other lines and was satisfied with the predictions by saying that ``\textit{I can trust its prediction a lot.}''

\subsection{Usage Scenario II: Credit Card Bill Prediction}
\label{Usage Scenario II}
\par
\begin{figure*}[h]
    \centering
     \vspace{-2mm}
    \includegraphics[width=\linewidth]{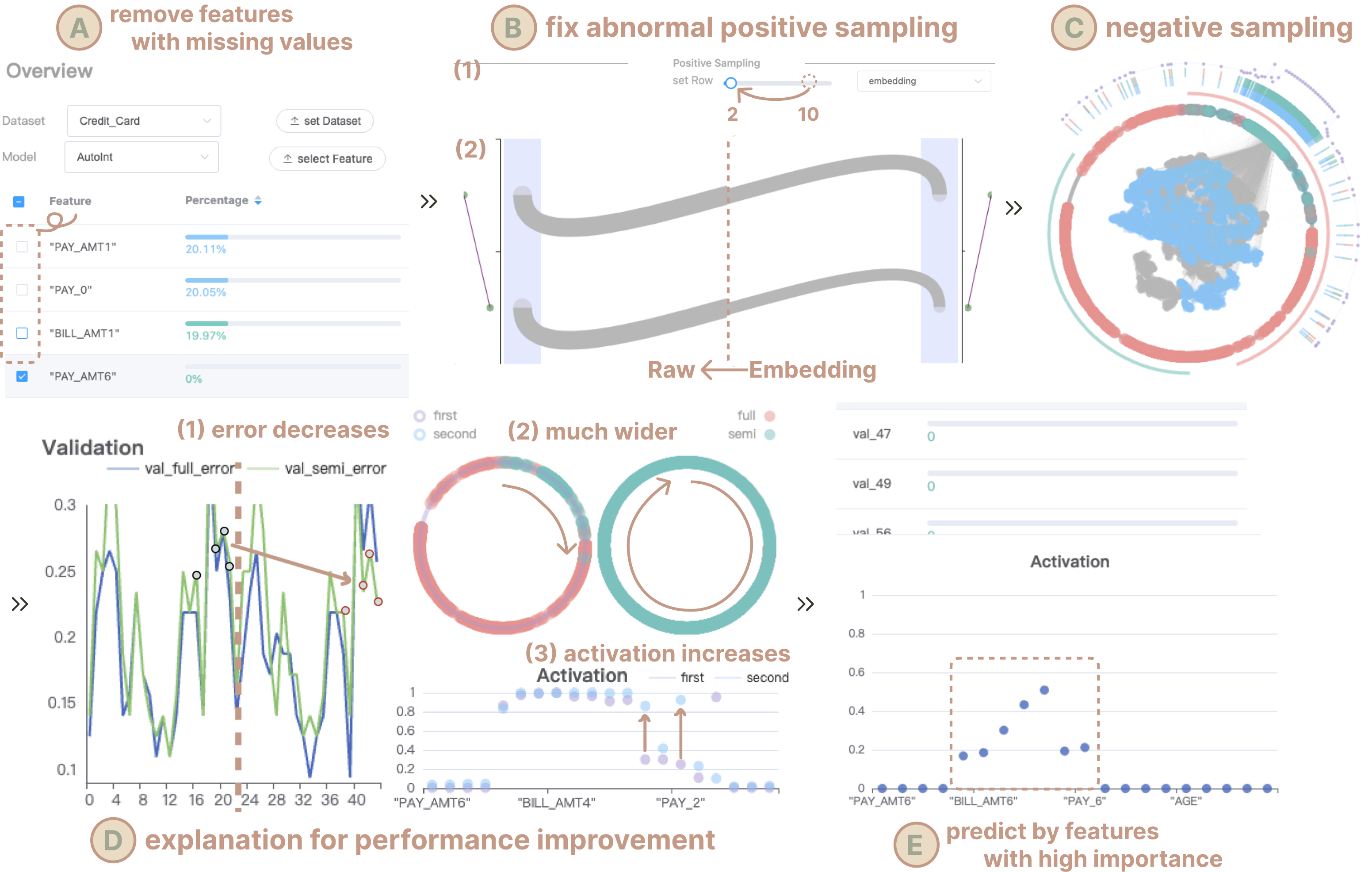}
        \vspace{-6mm}
    \caption{Usage scenario \change{II}. The notations from A to G refer to the steps in usage scenario II, in which we use visual tips to highlight the key operations or key discovery.}
    \label{fig:case2}
            \vspace{-6mm}
\end{figure*}

 We use another credit card dataset~\cite{CreditCard} with a model named \textit{AutoInt}~\cite{song2019autoint} \change{ to verify the efficacy of \textit{CIVis} in achieving a promising alignment}, which is recommended by E3. The credit card dataset includes a feature dimension of $25$, and the task is to predict whether the customer will pay the bill next month. \textit{AutoInt} uses unordered features and is designed for classification. We still divide the training/test set at the same scale to be consistent with the usage scenario I. However, the dataset has no missing data, so we randomly ($20\%$) remove some values of the $3$ dimensions (\textit{PAY\_AMT1}, \textit{PAY\_0}, and \textit{BILL\_AMT1}) (\autoref{fig:case2}(A)).

\par \textbf{Anomalous Positive Sampling}. The \textit{Set Row} function is adjusted to $2$, and the positive sampling representation is set to \textit{embedding} by default due to binary classification (\autoref{fig:case2}(B)(1)). In \autoref{fig:case2}(A), E3 first deselected the $3$ dimensions with missing values and clicked the \textit{select Feature} button, assuming no missing data during the inference process. The positive and negative sampling views are then updated accordingly, and the guidance view proceeds to \textit{Sampling}. In the case of binary classification, two thin rows appear in the positive sampling view (\change{right side in \autoref{fig:case2}(B)(2)}). \change{The thin links imply that the positive sampling is highly homogeneous.} ``\textit{Model collapse occurs or deteriorates when most semi-data are trained closer to the few full data}''. E3 pointed out and considered that the current positive sampling is anomalous. Thus, E3 conjectured that one of the following three scenarios occurred: 1) Model collapse occurs in the semi-model (i.e., most full data corresponds to a few semi-data in terms of its embeddings); 2) Model collapse occurs in the full model (i.e., most semi-data corresponds to a few full data in terms of its embeddings); 3) No model collapse occurs (i.e., semi-data and full data have few correspondences on their raw data.

\par To validate \change{these hypotheses}, E3 first switched to an alternative \change{representation \textit{raw} for positive sampling}, applying the positive sampling algorithm to the raw data instead of the embedding. Two parallel curves with thicker widths are shown (\change{left side} in \autoref{fig:case2}(B)(2)). Then, E3 switched to the negative sampling view to diagnose the model collapse from the perspective of uniformity. Most of the green dots have a narrower aggregation range than the red dots, indicating semi-data and full data, respectively (\autoref{fig:case2}(D)(2)). E3 attributed the anomalous positive sampling problem to the semi-model, and the resulting embeddings are not sufficiently different from each other. \change{Thereafter, he was satisfied with the \textit{raw} option for positive sampling and clicked the \textit{set Positive} button.}

\par \textbf{Alignment in the Prediction.} Based on the narrow green and wide red shown in the negative sampling view (\change{\autoref{fig:case2}(C)}), E3 sampled $90\%$ of the semi-data and $20\%$ of the full data by lassoing the points and applying the \textit{random} strategy. He then dragged the slider in the \textit{random} strategy to sample $90\%$ of the selected samples and clicked the \textit{set Negative} after \textit{m} is dragged to $0.99$ to induce the hardness of negative sampling and avoid potential model collapse. The guidance view proceeded to \textit{Train}. Before training, he clicked the \textit{add} button in the Logs subview to save the initialized model and \change{started the $0$-epoch training to trigger the update of the comparison view. The current distribution in the negative sampling view is preserved in the comparison view to allow the comparison of the results before and after training.} In the training view, he set \textit{Temperature} to $0.07$, \textit{Epoch} to $20$, and \textit{M} to $0.1$ and then clicked the \textit{train} button to start training. After the training is completed, the validation curve shows a clear and promising result. With a $22$ split on the x-axis, the blue curve is lower than the green curve (i.e., the loss given by the semi- and the full models), but they share a similar shape (left panel~\autoref{fig:case2}(D)(1)). Then, the blue curve dives deeper, and the green curve follows and in some cases exceeds (right panel in~\autoref{fig:case2}(D)(1)). This observation suggests that the training phase brings the semi-model in line with the full model.

\par E3 moved to the automatically updated comparison view to \change{determine} what brings the performance improvement. In the embedding distribution, the green points follow the distribution of the red points scattered on the circle. ``\textit{I am surprised that the semi-model could exactly mimic the distribution of the full model}'', said E3. In addition, he found that the blue dots raise above the purple dots in the activation of the comparison view (\autoref{fig:case2}(D)(3)). ``\textit{I think that the training distributions of the semi-model are caused by using more feature dimensions to distinguish from each other}'', said E3. Thereafter, he clicked the \textit{add} button again to test the overall classification error, which decreased \change{from 0.2195 to 0.2126}. E3 was satisfied with the trained model and the reasons for the improvement. He finally switched to the inference view to remove his last concern about overfitting. ``\textit{In binary classification, the accuracy is still high and the percentage is higher when the model predicts only one label}'', said E3. In this case, the dependence of activation on feature values is small and very low. He clicked on the \textit{infer} button and then used the rows with predicted values of zero or one to check out the feature importance (\autoref{fig:case2}(E)). The model makes predictions based on the features with high importance, so the display of all these features increases E3's confidence in the model's performance (i.e., accuracy).

\subsection{Quantitative Experiment}

\par We performed a quantitative experiment comparing \textit{CIVis} with classic data imputation algorithms. Following what we found in the first usage scenario of regression, we looked for the best ratio between the full data and the semi-data added to the negative collection and used the best semi-model found. We experimented with house price prediction using a test set ($10\%$). Few imputation algorithms can be directly utilized \change{because the features may be numbers or strings}. Accordingly, we chose \textit{KNN} and \textit{Most frequent~\cite{Imputation}}. We first imputed the missing values and then used the trained full model to predict the house prices. In the validation set, the MSE of the trained full model is $0.1123$, while the MSE of our best semi-model is $0.11895$. The results of the test data show that our semi-model achieved the best performance (\autoref{tab:exp}). In addition, we trained a semi-model without the help of \textit{CIVis} \change{to fairly demonstrate the performance due to \textit{CIVis}}, which is shown as a pure semi-model in~\autoref{tab:exp}. The \textit{Pure semi-model} is the worst of the four models ($0.207861$). Thus, we consider that for \textit{CIVis} there is a $62.32\%$ performance improvement.

\begin{table}[h]
\caption{Comparison with two imputation algorithms.}
\label{tab:exp}
\centering
\begin{tabular}{c|c|c|c|c} 
\hline \hline
    & KNN  & Most frequent & CIVis & Pure semi-model    \\ 
\hline \hline
MSE & 0.136286 & 0.136439      & \textbf{0.078315} & 0.207861 \\
ACC & 0.8107 & \textbf{0.8150} & 0.796 & 0.8013 \\
AUC & 0.6401 & \textbf{0.6407} & 0.612 & 0.5869 \\
\hline
\end{tabular}
\end{table}

\par We found the best full and semi-models according to the second usage scenario \change{following a similar procedure} and evaluated their performance on the test set ($10\%$). The differences lie in the dataset (i.e., the credit card dataset for classification) and the measurements (i.e., accuracy and AUC), which are common in classification. In \autoref{tab:exp}, the \textit{most frequent} achieves the best in both measures. However, the results of \textit{CIVis} are comparable, with its accuracy dropping by only $2.33\%$ and its AUC dropping by $4.78\%$. \change{In comparison with} \textit{pure semi-model}, \textit{CIVis} brought an accuracy loss of $0.66\%$ ($0.8013\rightarrow 0.796$), but the AUC improved by $4.28\%$ ($0.5869\rightarrow 0.612$).

\par The different performances in the two scenarios lead to the following observations: \textbf{1) CL yields different improvements depending on the task.} CL pushes negative pairs and pulls positive pairs, a necessary benefit for latent representation learning, but the impact on downstream tasks is uncertain. For example, the regression task in \change{usage scenario I} gains a surprising boost from the properties of CL because the prediction of the floating-point numbers requires a subtle difference from \change{pushed} embeddings. By contrast, the classification task requires critical distinctions, wherein the detailed refinement may be ignored, while the \change{pushed} embedding moderates salient features and classification accuracy. Meanwhile, the harmful \change{pushing} prevents overfitting and increases the AUC, acting like a regularization term. \textbf{2) CL may conflict with the \change{selected} model.} \change{In the classification scenario, the selected \textit{AutoInt} is originally powerful because it induces the dynamic feature weights from the Attention module.} The best performance is obtained by \textit{Most frequent}, which implies that the involvement of \textit{CIVis} compromises the effectiveness of \textit{AutoInt}. The CL approach may induce negative effects when the model is powerful enough. \textbf{3) Results of \textit{KNN} and \textit{Most frequent} depend on the missing data.} \textit{KNN} and \textit{Most frequent} have strong assumptions on the distribution of missing values. When we perform another random removal (discussed in~\autoref{Usage Scenario II}), their results are different in terms of accuracy and AUC, while \textit{CIVis} and \textit{Pure semi-model} are consistent. Thus, \textit{CIVis} is the best solution when stability is important enough to sacrifice reasonable accuracy from the perspective of missing data.

\subsection{Expert Interview}
\par We conducted a one-hour semi-structured interview with our collaborating experts (E1-E3) to assess whether our approach helped them in inducing domain knowledge in the absence of data.

\par \textbf{System capabilities and learning curve.} All experts appreciated the ability of \textit{CIVis} to support interactive configurations to make predictions by using full and semi-data. The experts also expressed satisfaction with the customized visual design and interaction of the system. We deliberately selected familiar visual metaphors from the CL domain to help the experts quickly become familiar with our visual encoding. In addition, we conducted a user-centered design process. After introducing the system, the experts could develop a customized exploration path.

\par \textbf{Scalability.} The experts pointed out the advantages and disadvantages of instance-level visualization. E1 argued that presenting raw data is important for users to understand examples of algorithmic output. However, E2 discussed his concerns about scalability. Modern open-source datasets are always large-scale, and the visual designs that render each instance pose serious performance problems, such as latency. E3 suggested moving our system to a high-performance platform, such as WebGL when we work with large-scale datasets.

\subsection{Qualitative User Study}
\par We conduct a user study to further evaluate the effectiveness of \textit{CIVis} in four areas: \textit{Informativeness}, \textit{Decision Making}, \textit{Visual Design and Interaction}, \textit{Usability and Perception}, following a four-layer taxonomy~\cite{weibelzahl2020evaluation}.

\par \textbf{Participants.} We recruited $12$ participants for the user study ($4$ females, $8$ males, $age_{mean} = 24.5$, $age_{sd} = 3.55$) via word-of-mouth, including students and practitioners from the major of Computer Science at ShanghaiTech University and Tencent. The participants all have at least $2$ years of ML experience. \change{Ten} of the participants are students, and the remaining two participants are ML practitioners from the company. In particular, four of the participants have experience or knowledge about CL, and we pay more attention to their ratings and comments. The study was conducted face-to-face, and the users participated in tutorials, experienced the system, performed tasks, and filled out questionnaires. We use $P1-P12$ to denote the participants, $M/F$ to denote the participants' gender as male/female, and $C/NC$ to denote the participants' experience with/without CL. For example, the $9^{th}$ participant is male, is $25$ years old, and has experience with CL, denoted by $P9$ ($M, 25, C$).

\par \textbf{Tasks.} In our user study, each participant was asked to complete a task in turn. The task had to be completed within $45$ min and had two purposes: 1) A CNN model is trained to infer house prices from the incomplete house values, and 2) CL is used to refine the performance of the semi-model to approach that of the full model. Given that some participants (i.e., students) are not real-world ML practitioners, we primarily examine the integrity and usability of the system design. For more discussion on system effectiveness, please refer to the two usage scenarios above.

\par \textbf{Procedure.} The entire process lasted \change{for} approximately $1$ h. Before we demonstrated our system, the participants were informed that their responses and feedback would be anonymously collected in a questionnaire. Then, we conducted a \modify{15-20} \change{min} tutorial session during which we demonstrated the complete workflow of \textit{CIVis} while explaining all the operations of each view. \change{Specifically}, the system tutorial showed how CL can help solve the problem of modeling observed data with missing values. When the participants felt ready, they were given the aforementioned tasks to complete individually. After completing the task, a questionnaire was distributed with questions on a $7$-point Likert scale, with $1$ representing ``strongly disagree'' and $7$ representing ``strongly agree'' (\autoref{tab:question}). The other comments that were not included in the questionnaire were also recorded for reference.

\begin{table}[h]
\caption{Assessment of \textit{CIVis} in terms of \textit{informativeness} (Q1-Q3), \textit{decision making} (Q4-Q6), \textit{visual design and interaction} (Q7-Q9), and \textit{usability and perception} (Q10-Q13).}
\label{tab:question}
\centering
\begin{tabular}{l|l} 
\toprule
\multicolumn{2}{l}{\textbf{Informativeness}}                                                                                                                                                                 \\ 
\hline
Q1  & \textcolor[rgb]{0.2,0.2,0.2}{The overall situation of missing data is easy to access.}                                                                                                        \\
Q2  & \begin{tabular}[c]{@{}l@{}}\textcolor[rgb]{0.2,0.2,0.2}{The information for sampling and learning process }\\\textcolor[rgb]{0.2,0.2,0.2}{of CL is detailed and rich.}\end{tabular}           \\
Q3  & \textcolor[rgb]{0.2,0.2,0.2}{\textit{CIVis} provides sufficient information of the workflow.}                                                                                                          \\ 
\hline
\multicolumn{2}{l}{\textbf{Decision Making}}                                                                                                                                                                 \\ 
\hline
Q4  & \begin{tabular}[c]{@{}l@{}}\textcolor[rgb]{0.2,0.2,0.2}{\textit{CIVis} facilitates inducing domain knowledge into the }\\\textcolor[rgb]{0.2,0.2,0.2}{process of contrastive learning.}\end{tabular}  \\
Q5  & \begin{tabular}[c]{@{}l@{}}\textcolor[rgb]{0.2,0.2,0.2}{\textit{CIVis} provides clues to understand and debug }\\\textcolor[rgb]{0.2,0.2,0.2}{contrastive learning during training.}\end{tabular}     \\
Q6  & \textcolor[rgb]{0.2,0.2,0.2}{\textit{CIVis} helps my decision-making during inference.}                                                                                                                \\ 
\hline
\multicolumn{2}{l}{\textbf{Visual Design and Interaction}}                                                                                                                                                  \\ 
\hline
Q7  & \textcolor[rgb]{0.2,0.2,0.2}{The workflow is formed reasonably and logically.}                                                                                                               \\
Q8  & \textcolor[rgb]{0.2,0.2,0.2}{The interaction facilitates the workflow.}                                                                                                                      \\
Q9  & \textcolor[rgb]{0.2,0.2,0.2}{The visual design is intuitive and easy to understand.}                                                                                                          \\ 
\hline
\multicolumn{2}{l}{\textbf{Usability and Perception}}                                                                                                                                                       \\ 
\hline
Q10 & \textcolor[rgb]{0.2,0.2,0.2}{The system can help me train an accurate model.}                                                                                                                \\
Q11 & \textcolor[rgb]{0.2,0.2,0.2}{I am willing to trust the results given by \textit{CIVis}.}                                                                                                          \\
Q12 & \textcolor[rgb]{0.2,0.2,0.2}{\textit{CIVis} is easy to learn and convenient to use.}                                                                                                                   \\
Q13 & \textcolor[rgb]{0.2,0.2,0.2}{I would like to recommend \textit{CIVis} to other scenarios.}                                                                                                             \\
\bottomrule
\end{tabular}
\vspace{-3mm}
\end{table}

\par
\begin{figure}[h]
    \vspace{-6mm}
    \centering
    \includegraphics[width=\linewidth]{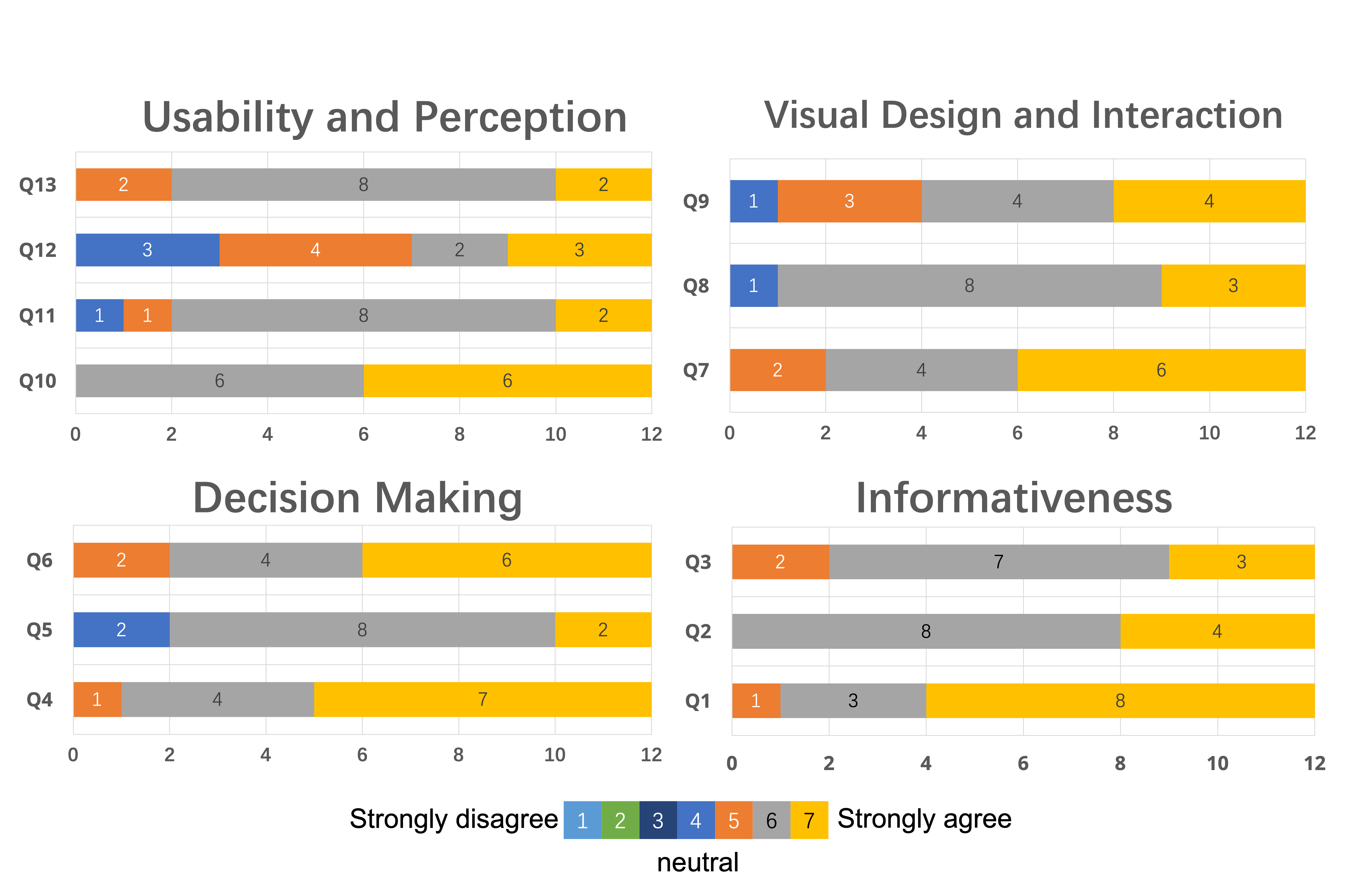}
    \vspace{-6mm}
    \caption{The results of the user study were divided into four areas. The average rate from high to low was $6.33$ for informativeness, $6.22$ for decision-making, \modify{$6.11$} for visual design and interaction, and \modify{$5.96$} for usability and perception.}
    \label{fig:user_study}
        \vspace{-6mm}
\end{figure}

 \textbf{Results.} The detailed statistics of the questionnaire are shown in \autoref{fig:user_study}. \textit{CIVis} was appreciated by most participants, \change{with} no ratings below $4$ and an overall average rating of \modify{6.14}. This evaluation result has several interesting features. First, we observed only $6$ or $7$ ratings in $Q2$ and $Q10$ with average rates of $6.33$ and $6.5$, \change{indicating} that the system design \change{satisfactorily} satisfies the main objective of training a transparent CL model (\textbf{R.2 and R.3}). ``\textit{Even though I know little about CL, I can understand the obvious benefit of CL in pushing the negative samples away and pulling the positive samples together}'', said $P2$ ($F, 23, NC$). Second, the higher average rates of informativeness and decision-making ($6.33$ and $6.22$) can cover all the requirements (\textbf{R.1-R.4}). This notion means that the implementation of \textit{CIVis} succeeded in achieving the blueprint we had at the beginning. ''\textit{I am impressed with the direct control of the embedding distribution}''. The $P5$ ($M, 23, C$) with CL experience praised the ability to induce domain knowledge into the model training. Third, $Q13$ had the lowest mean rate ($5.25$) and the lowest mean rate for usability and perception ($5.92$). We discuss these two observations together because this question set also includes the highly rated question $Q10$. Accordingly, we could infer that it is the relatively low rating of $Q13$ that leads to the relatively low rating of usability and perception. Although the concept of CL and its benefits for solving the problem of modeling observed data with missing values took some time to understand, participants could learn and use \textit{CIVis}. When we asked the participants about their reasons for the learning curves, their focus reflected on understanding the background of CL and figuring out our innovations in using CL to solve the problem of modeling observed data with missing values. ``\textit{I think my difficulty is figuring out how to use CL to improve models with missing features, not the usability of the system}'', said $P3$ ($M, 23, NC$). This \change{circumstance} inspired us to count participants with or without CL experience, who had mean ratings of $6.33$ and \modify{$6.05$}, respectively. Thus, we conclude that the learning curve of CL and its application to modeling observed data with missing values led to relatively low ratings in terms of usability and perception. Finally, the visual design and interaction in \textit{CIVis} received an average rate of \modify{$6.11$}, which is a bit lower than the overall average (\modify{$6.14$}). ''\textit{The visualization is easy to understand, but there is still room for improvement, for example, in the positive sampling view, the highlight should be obvious when I click on a circle}'', commented $P12$ ($M, 33, NC$) with previous experience using visual analytics systems.

\section{Discussion and Limitations}
\par In this section, we discuss the high-level synthetics, the design implications, and the limitations of our approach.

\par \textbf{Different Path to Resolving Modeling Observed Data with Missing Values}. Current approaches to improving ML modeling performance generally prioritize dealing with missing data, which can be problematic as imputation techniques estimate missing values more or less (\autoref{sec:missing}). While ML has enabled increasingly accurate missing value estimation, this approach can never perfectly replicate the true distribution. Sarma et al.~\cite{sarma2022evaluating} addressed the visualization of missing data uncertainty through scatterplots and analyzed missingness patterns, and conducted a crowdsourcing study to assess the impact of visual representation on reasoning. However, this work remains confined to automatic imputation. Our approach involves removing missing features and training prediction models using fewer features to avoid the need for imputation. Furthermore, \textit{CIVis} and imputations are influenced by domain knowledge in different ways. While \textit{CIVis} is guided by CL theory, imputations depend on the uncertainty of missing data. We believe that our method can help facilitate the modeling of observed data with missing values and provide a novel approach for integrating domain knowledge into the model building process.

\par \modify{\textbf{Contribution Over Previous Studies of CL.} The scope of prior works on design principles for CL is limited to the image domain. These studies suggest two key principles: (1) \textbf{alignment} and \textbf{uniformity}, which entails ensuring that the distance distributions of sampled embeddings are close to their positive pairs and far from their negative pairs~\cite{wang2020understanding,tian2020makes}, and (2) \textbf{hardness}, which refers to the existence of a suitable level of difficulty for positive pairs~\cite{grill2020bootstrap,zhu2021improving,tian2020makes,zhang2022does}. We have incorporated these principles into our system, and their validity has been confirmed on a dataset with another modality of the unstructured features typically seen in recommendation systems, as shown in two usage scenarios. Our work extends the existing theoretical analysis to a broader domain, thus highlighting the general applicability of CL. However, the two principles outlined above are somewhat contradictory in that induced hardness reduces alignment. This phenomenon has been previously presented in the studies~\cite{zhang2022does}, which analyzed how state-of-the-art CL models are implemented, a model-dependent explanation, but the theoretical solution for the optimal trade-off between alignment and hardness remains unsolved. Our system, \textit{CIVis}, allows users to explore potential solutions for this problem by adjusting the sampling ratio between semi- and full data and can serve as a source of inspiration for future research in this area.}


\par \textbf{Implication for Explainable ML.} \textit{CIVis} introduces the use of CL for model debugging, enabled by visual analytics. The CL-based debugging method offers several advantages. First, it is agnostic to the specific deep learning model, as demonstrated by its successful application in two different usage scenarios. Second, the circular visualization of data distributions represents a novel approach that can uncover the reasons behind both positive and negative model performance in a post hoc explanation. Third, runtime metrics for sampled pairs facilitate the rapid detection of anomalies during the training process, providing an advantage over post hoc techniques that require longer waiting periods. Previous research efforts in the field of model interpretability, as discussed in~\autoref{sec:explain}, have incorporated one or two of the aforementioned features, but few have integrated all of them. Thus, the proposed CL-based debugging methodology has the potential to motivate further research endeavors that exploit its features and benefit a wider range of machine learning models through its multi-phased interpretability, both during and after the training phase.

\par \textbf{Limitation.} This study has a few limitations that should be taken into consideration. First, the scalability of \textit{CIVis} may be an issue when a large number of nodes are involved in the projection. A potential solution to this issue is to eliminate invisible points (i.e., points that are not visible due to overlapping) and select samples based on their positions. Second, while experts may be able to train an acceptable ML model using \textit{CIVis}, the learning curve at the beginning is steep, which may require users to have a background in ML. Third, when most or key features are missing, \textit{CIVis} fails to perform, rendering it ineffective. In such cases, no methods may be suitable for the task at hand.

\section{Conclusion and Future Work}
\par This study presents a framework based on CL that is capable of modeling observed data with missing values in the ML pipeline. Our proposed framework utilizes both full data and semi-data, without any form of imputation. In addition, we introduce a visual analytics system called \textit{CIVis}, which integrates the CL-based framework to assist experts in leveraging their domain knowledge and engaging in co-adaptive decision-making. The effectiveness of our approach has been validated through two usage scenarios, a quantitative experiment, expert interviews, and a qualitative user study. As a future direction, we plan to extend our CL-based framework to other models and downstream tasks.

\bibliographystyle{IEEEtran}
\bibliography{template.bib}

\begin{thebibliography}{10}
\providecommand{\url}[1]{#1}
\csname url@samestyle\endcsname
\providecommand{\newblock}{\relax}
\providecommand{\bibinfo}[2]{#2}
\providecommand{\BIBentrySTDinterwordspacing}{\spaceskip=0pt\relax}
\providecommand{\BIBentryALTinterwordstretchfactor}{4}
\providecommand{\BIBentryALTinterwordspacing}{\spaceskip=\fontdimen2\font plus
\BIBentryALTinterwordstretchfactor\fontdimen3\font minus
  \fontdimen4\font\relax}
\providecommand{\BIBforeignlanguage}[2]{{%
\expandafter\ifx\csname l@#1\endcsname\relax
\typeout{** WARNING: IEEEtran.bst: No hyphenation pattern has been}%
\typeout{** loaded for the language `#1'. Using the pattern for}%
\typeout{** the default language instead.}%
\else
\language=\csname l@#1\endcsname
\fi
#2}}
\providecommand{\BIBdecl}{\relax}
\BIBdecl

\bibitem{rubin2004multiple}
D.~B. Rubin, \emph{Multiple imputation for nonresponse in surveys}.\hskip 1em
  plus 0.5em minus 0.4em\relax John Wiley \& Sons, 2004, vol.~81.

\bibitem{van2011mice}
S.~Van~Buuren and K.~Groothuis-Oudshoorn, ``Mice: Multivariate imputation by
  chained equations in {R},'' \emph{Journal of Statistical Software}, vol.~45,
  pp. 1--67, 2011.

\bibitem{honaker2011amelia}
J.~Honaker, G.~King, and M.~Blackwell, ``Amelia {II}: A program for missing
  data,'' \emph{Journal of Statistical Software}, vol.~45, no.~1, pp. 1--47,
  2011.

\bibitem{hastie2015matrix}
T.~Hastie, R.~Mazumder, J.~D. Lee, and R.~Zadeh, ``Matrix completion and
  low-rank {SVD} via fast alternating least squares,'' \emph{The Journal of
  Machine Learning Research}, vol.~16, no.~1, pp. 3367--3402, 2015.

\bibitem{xia2017adjusted}
J.~Xia, S.~Zhang, G.~Cai, L.~Li, Q.~Pan, J.~Yan, and G.~Ning, ``Adjusted weight
  voting algorithm for random forests in handling missing values,''
  \emph{Pattern Recognition}, vol.~69, pp. 52--60, 2017.

\bibitem{smieja2018processing}
M.~Smieja, {\L}.~Struski, J.~Tabor, B.~Zieli{\'n}ski, and P.~Spurek,
  ``Processing of missing data by neural networks,'' \emph{arXiv preprint
  arXiv:1805.07405}, 2018.

\bibitem{you2020handling}
J.~You, X.~Ma, D.~Y. Ding, M.~Kochenderfer, and J.~Leskovec, ``Handling missing
  data with graph representation learning,'' \emph{arXiv preprint
  arXiv:2010.16418}, 2020.

\bibitem{sarma2022evaluating}
A.~Sarma, S.~Guo, J.~Hoffswell, R.~Rossi, F.~Du, E.~Koh, and M.~Kay,
  ``Evaluating the use of uncertainty visualisations for imputations of data
  missing at random in scatterplots,'' \emph{IEEE Transactions on Visualization
  and Computer Graphics}, vol.~29, no.~1, pp. 602--612, 2022.

\bibitem{liu2021self}
X.~Liu, F.~Zhang, Z.~Hou, L.~Mian, Z.~Wang, J.~Zhang, and J.~Tang,
  ``Self-supervised learning: Generative or contrastive,'' \emph{IEEE
  Transactions on Knowledge and Data Engineering}, vol.~35, no.~1, pp.
  857--876, 2021.

\bibitem{gillies2016human}
M.~Gillies, R.~Fiebrink, A.~Tanaka, J.~Garcia, F.~Bevilacqua, A.~Heloir,
  F.~Nunnari, W.~Mackay, S.~Amershi, B.~Lee, N.~d'Alessandro, J.~Tilmanne,
  T.~Kulesza, and B.~Caramiaux, ``Human-centred machine learning,'' in
  \emph{CHI conference extended abstracts on human factors in computing
  systems}, 2016, pp. 3558--3565.

\bibitem{schroff2015facenet}
F.~Schroff, D.~Kalenichenko, and J.~Philbin, ``Facenet: A unified embedding for
  face recognition and clustering,'' in \emph{IEEE/CVF Conference on Computer
  Vision and Pattern Recognition}, 2015, pp. 815--823.

\bibitem{suh2019stochastic}
Y.~Suh, B.~Han, W.~Kim, and K.~M. Lee, ``Stochastic class-based hard example
  mining for deep metric learning,'' in \emph{IEEE/CVF Conference on Computer
  Vision and Pattern Recognition}, 2019, pp. 7251--7259.

\bibitem{robinson2020contrastive}
J.~Robinson, C.-Y. Chuang, S.~Sra, and S.~Jegelka, ``Contrastive learning with
  hard negative samples,'' \emph{arXiv preprint arXiv:2010.04592}, 2020.

\bibitem{he2020momentum}
K.~He, H.~Fan, Y.~Wu, S.~Xie, and R.~Girshick, ``Momentum contrast for
  unsupervised visual representation learning,'' in \emph{IEEE/CVF Conference
  on Computer Vision and Pattern Recognition}, 2020, pp. 9729--9738.

\bibitem{li2021model}
Q.~Li, B.~He, and D.~Song, ``Model-contrastive federated learning,'' in
  \emph{IEEE/CVF Conference on Computer Vision and Pattern Recognition}, 2021,
  pp. 10\,713--10\,722.

\bibitem{selvaraju2017grad}
R.~R. Selvaraju, M.~Cogswell, A.~Das, R.~Vedantam, D.~Parikh, and D.~Batra,
  ``Grad-cam: Visual explanations from deep networks via gradient-based
  localization,'' in \emph{IEEE/CVF International Conference on Computer
  Vision}, 2017, pp. 618--626.

\bibitem{xuan2020improved}
H.~Xuan, A.~Stylianou, and R.~Pless, ``Improved embeddings with easy positive
  triplet mining,'' in \emph{IEEE/CVF Winter Conference on Applications of
  Computer Vision}, 2020, pp. 2474--2482.

\bibitem{grill2020bootstrap}
J.-B. Grill, F.~Strub, F.~Altch\'{e}, C.~Tallec, P.~H. Richemond,
  E.~Buchatskaya, C.~Doersch, B.~A. Pires, Z.~D. Guo, M.~G. Azar, B.~Piot,
  K.~Kavukcuoglu, R.~Munos, and M.~Valko, ``Bootstrap your own latent a new
  approach to self-supervised learning,'' in \emph{International Conference on
  Neural Information Processing Systems}, ser. NIPS'20.\hskip 1em plus 0.5em
  minus 0.4em\relax Red Hook, NY, USA: Curran Associates Inc., 2020.

\bibitem{kim2004reuse}
K.-Y. Kim, B.-J. Kim, and G.-S. Yi, ``Reuse of imputed data in microarray
  analysis increases imputation efficiency,'' \emph{BMC Bioinformatics},
  vol.~5, no.~1, pp. 1--9, 2004.

\bibitem{gondara2017multiple}
L.~Gondara and K.~Wang, ``Multiple imputation using deep denoising
  autoencoders,'' \emph{arXiv preprint arXiv:1705.02737}, 2017.

\bibitem{yoon2018gain}
J.~Yoon, J.~Jordon, and M.~Schaar, ``Gain: Missing data imputation using
  generative adversarial nets,'' in \emph{International Conference on Machine
  Learning}.\hskip 1em plus 0.5em minus 0.4em\relax PMLR, 2018, pp. 5689--5698.

\bibitem{chen2020simple}
T.~Chen, S.~Kornblith, M.~Norouzi, and G.~Hinton, ``A simple framework for
  contrastive learning of visual representations,'' in \emph{International
  Conference on Machine Learning}.\hskip 1em plus 0.5em minus 0.4em\relax PMLR,
  2020, pp. 1597--1607.

\bibitem{zhu2021improving}
R.~Zhu, B.~Zhao, J.~Liu, Z.~Sun, and C.~W. Chen, ``Improving contrastive
  learning by visualizing feature transformation,'' in \emph{IEEE/CVF
  International Conference on Computer Vision}, 2021, pp. 10\,306--10\,315.

\bibitem{yao2021g}
L.~Yao, R.~Pi, H.~Xu, W.~Zhang, Z.~Li, and T.~Zhang, ``G-detkd: Towards general
  distillation framework for object detectors via contrastive and
  semantic-guided feature imitation,'' in \emph{IEEE/CVF International
  Conference on Computer Vision}, 2021, pp. 3591--3600.

\bibitem{zhou2021contrastive}
C.~Zhou, J.~Ma, J.~Zhang, J.~Zhou, and H.~Yang, ``Contrastive learning for
  debiased candidate generation in large-scale recommender systems,'' in
  \emph{ACM SIGKDD Conference on Knowledge Discovery \& Data Mining}, 2021, pp.
  3985--3995.

\bibitem{oord2018representation}
A.~v.~d. Oord, Y.~Li, and O.~Vinyals, ``Representation learning with
  contrastive predictive coding,'' \emph{arXiv preprint arXiv:1807.03748},
  2018.

\bibitem{bors2018visual}
C.~Bors, T.~Gschwandtner, S.~Kriglstein, S.~Miksch, and M.~Pohl, ``Visual
  interactive creation, customization, and analysis of data quality metrics,''
  \emph{Journal of Data and Information Quality (JDIQ)}, vol.~10, no.~1, pp.
  1--26, 2018.

\bibitem{xiang2019interactive}
S.~Xiang, X.~Ye, J.~Xia, J.~Wu, Y.~Chen, and S.~Liu, ``Interactive correction
  of mislabeled training data,'' in \emph{IEEE Conference on Visual Analytics
  Science and Technology (VAST)}.\hskip 1em plus 0.5em minus 0.4em\relax IEEE,
  2019, pp. 57--68.

\bibitem{kandel2012profiler}
S.~Kandel, R.~Parikh, A.~Paepcke, J.~M. Hellerstein, and J.~Heer, ``Profiler:
  Integrated statistical analysis and visualization for data quality
  assessment,'' in \emph{International Working Conference on Advanced Visual
  Interfaces}, 2012, pp. 547--554.

\bibitem{angelini2022visual}
M.~Angelini, C.~Daraio, and L.~Urban, ``A visual analytics approach for the
  assessment of information quality of performance models—a software
  review,'' \emph{Scientometrics}, vol. 127, no.~12, pp. 6827--6853, 2022.

\bibitem{liu2016towards}
M.~Liu, J.~Shi, Z.~Li, C.~Li, J.~Zhu, and S.~Liu, ``Towards better analysis of
  deep convolutional neural networks,'' \emph{IEEE Transactions on
  Visualization and Computer Graphics}, vol.~23, no.~1, pp. 91--100, 2016.

\bibitem{ming2017understanding}
Y.~Ming, S.~Cao, R.~Zhang, Z.~Li, Y.~Chen, Y.~Song, and H.~Qu, ``Understanding
  hidden memories of recurrent neural networks,'' in \emph{IEEE Conference on
  Visual Analytics Science and Technology (VAST)}.\hskip 1em plus 0.5em minus
  0.4em\relax IEEE, 2017, pp. 13--24.

\bibitem{derose2020attention}
J.~F. DeRose, J.~Wang, and M.~Berger, ``Attention flows: Analyzing and
  comparing attention mechanisms in language models,'' \emph{IEEE Transactions
  on Visualization and Computer Graphics}, vol.~27, no.~2, pp. 1160--1170,
  2020.

\bibitem{jin2022gnnlens}
Z.~Jin, Y.~Wang, Q.~Wang, Y.~Ming, T.~Ma, and H.~Qu, ``Gnnlens: A visual
  analytics approach for prediction error diagnosis of graph neural networks,''
  \emph{IEEE Transactions on Visualization and Computer Graphics}, vol.~29,
  no.~6, pp. 3024--3038, 2023.

\bibitem{gou2020vatld}
L.~Gou, L.~Zou, N.~Li, M.~Hofmann, A.~K. Shekar, A.~Wendt, and L.~Ren, ``Vatld:
  A visual analytics system to assess, understand and improve traffic light
  detection,'' \emph{IEEE Transactions on Visualization and Computer Graphics},
  vol.~27, no.~2, pp. 261--271, 2020.

\bibitem{la2023state}
B.~La~Rosa, G.~Blasilli, R.~Bourqui, D.~Auber, G.~Santucci, R.~Capobianco,
  E.~Bertini, R.~Giot, and M.~Angelini, ``State of the art of visual analytics
  for explainable deep learning,'' in \emph{Computer Graphics Forum}, vol.~42,
  no.~1.\hskip 1em plus 0.5em minus 0.4em\relax Wiley Online Library, 2023, pp.
  319--355.

\bibitem{liu2017visual}
S.~Liu, J.~Xiao, J.~Liu, X.~Wang, J.~Wu, and J.~Zhu, ``Visual diagnosis of tree
  boosting methods,'' \emph{IEEE Transactions on Visualization and Computer
  Graphics}, vol.~24, no.~1, pp. 163--173, 2017.

\bibitem{kahng2018gan}
M.~Kahng, N.~Thorat, D.~H. Chau, F.~B. Vi{\'e}gas, and M.~Wattenberg, ``Gan
  lab: Understanding complex deep generative models using interactive visual
  experimentation,'' \emph{IEEE Transactions on Visualization and Computer
  Graphics}, vol.~25, no.~1, pp. 310--320, 2018.

\bibitem{wang2018dqnviz}
J.~Wang, L.~Gou, H.-W. Shen, and H.~Yang, ``Dqnviz: A visual analytics approach
  to understand deep q-networks,'' \emph{IEEE Transactions on Visualization and
  Computer Graphics}, vol.~25, no.~1, pp. 288--298, 2018.

\bibitem{turkay2018progressive}
C.~Turkay, N.~Pezzotti, C.~Binnig, H.~Strobelt, B.~Hammer, D.~A. Keim, J.-D.
  Fekete, T.~Palpanas, Y.~Wang, and F.~Rusu, ``Progressive data science:
  Potential and challenges,'' \emph{arXiv preprint arXiv:1812.08032}, 2018.

\bibitem{echihabi2022pros}
K.~Echihabi, T.~Tsandilas, A.~Gogolou, A.~Bezerianos, and T.~Palpanas, ``Pros:
  Data series progressive k-{NN} similarity search and classification with
  probabilistic quality guarantees,'' \emph{The VLDB Journal}, pp. 1--27, 2022.

\bibitem{zhang2022does}
C.~Zhang, K.~Zhang, C.~Zhang, T.~X. Pham, C.~D. Yoo, and I.~S. Kweon, ``How
  does simsiam avoid collapse without negative samples? a unified understanding
  with self-supervised contrastive learning,'' \emph{arXiv preprint
  arXiv:2203.16262}, 2022.

\bibitem{wang2020understanding}
T.~Wang and P.~Isola, ``Understanding contrastive representation learning
  through alignment and uniformity on the hypersphere,'' in \emph{International
  Conference on Machine Learning}.\hskip 1em plus 0.5em minus 0.4em\relax PMLR,
  2020, pp. 9929--9939.

\bibitem{wu2018unsupervised}
Z.~Wu, Y.~Xiong, S.~X. Yu, and D.~Lin, ``Unsupervised feature learning via
  non-parametric instance discrimination,'' in \emph{IEEE/CVF Conference on
  Computer Vision and Pattern Recognition}, 2018, pp. 3733--3742.

\bibitem{tian2020makes}
Y.~Tian, C.~Sun, B.~Poole, D.~Krishnan, C.~Schmid, and P.~Isola, ``What makes
  for good views for contrastive learning?'' \emph{Advances in Neural
  Information Processing Systems}, vol.~33, pp. 6827--6839, 2020.

\bibitem{shorten2019survey}
C.~Shorten and T.~M. Khoshgoftaar, ``A survey on image data augmentation for
  deep learning,'' \emph{Journal of Big Data}, vol.~6, no.~1, pp. 1--48, 2019.

\bibitem{feng2021survey}
S.~Y. Feng, V.~Gangal, J.~Wei, S.~Chandar, S.~Vosoughi, T.~Mitamura, and
  E.~Hovy, ``A survey of data augmentation approaches for {NLP},'' \emph{arXiv
  preprint arXiv:2105.03075}, 2021.

\bibitem{you2020graph}
Y.~You, T.~Chen, Y.~Sui, T.~Chen, Z.~Wang, and Y.~Shen, ``Graph contrastive
  learning with augmentations,'' \emph{Advances in Neural Information
  Processing Systems}, vol.~33, pp. 5812--5823, 2020.

\bibitem{liu2017sphereface}
W.~Liu, Y.~Wen, Z.~Yu, M.~Li, B.~Raj, and L.~Song, ``Sphereface: Deep
  hypersphere embedding for face recognition,'' in \emph{IEEE/CVF Conference on
  Computer Vision and Pattern Recognition}, 2017, pp. 212--220.

\bibitem{van2008visualizing}
L.~van~der Maaten and G.~Hinton, ``Visualizing data using t-{SNE},''
  \emph{Journal of Machine Learning Research}, vol.~9, no.~86, pp. 2579--2605,
  2008.

\bibitem{riehmann2005interactive}
P.~Riehmann, M.~Hanfler, and B.~Froehlich, ``Interactive sankey diagrams,'' in
  \emph{IEEE Symposium on Information Visualization (INFOVIS).}\hskip 1em plus
  0.5em minus 0.4em\relax IEEE, 2005, pp. 233--240.

\bibitem{wells1990longman}
J.~C. Wells and T.~T. Hung, ``Longman pronunciation dictionary,'' \emph{RELC
  Journal}, vol.~21, no.~2, pp. 95--97, 1990.

\bibitem{ceneda2019review}
D.~Ceneda, T.~Gschwandtner, and S.~Miksch, ``A review of guidance approaches in
  visual data analysis: A multifocal perspective,'' in \emph{Computer Graphics
  Forum}, vol.~38, no.~3.\hskip 1em plus 0.5em minus 0.4em\relax Wiley Online
  Library, 2019, pp. 861--879.

\bibitem{isenberg2013systematic}
T.~Isenberg, P.~Isenberg, J.~Chen, M.~Sedlmair, and T.~M{\"o}ller, ``A
  systematic review on the practice of evaluating visualization,'' \emph{IEEE
  Transactions on Visualization and Computer Graphics}, vol.~19, no.~12, pp.
  2818--2827, 2013.

\bibitem{HousePrice}
``House price,''
  \url{https://www.kaggle.com/c/house-prices-advanced-regression-techniques/overview/description},
  accessed: 2021-10-31.

\bibitem{mckinney2011pandas}
W.~McKinney, ``Pandas: A foundational {P}ython library for data analysis and
  statistics,'' \emph{Python for High Performance and Scientific Computing},
  vol.~14, no.~9, pp. 1--9, 2011.

\bibitem{CreditCard}
``Credit card,''
  \url{https://www.kaggle.com/datasets/uciml/default-of-credit-card-clients-dataset},
  accessed: 2022-08-12.

\bibitem{song2019autoint}
W.~Song, C.~Shi, Z.~Xiao, Z.~Duan, Y.~Xu, M.~Zhang, and J.~Tang, ``Autoint:
  Automatic feature interaction learning via self-attentive neural networks,''
  in \emph{ACM International Conference on Information and Knowledge
  Management}, 2019, pp. 1161--1170.

\bibitem{Imputation}
``Imputation,''
  \url{https://scikit-learn.org/stable/modules/classes.html\#module-sklearn.impute},
  accessed: 2021-12-3.

\bibitem{weibelzahl2020evaluation}
S.~Weibelzahl, A.~Paramythis, and J.~Masthoff, ``Evaluation of adaptive
  systems,'' in \emph{ACM Conference on User Modeling, Adaptation and
  Personalization}, 2020, pp. 394--395.

\end{thebibliography}

%
\vspace{-12mm}
\begin{IEEEbiography}[{\includegraphics[width=1in,height=1.25in,clip,keepaspectratio]{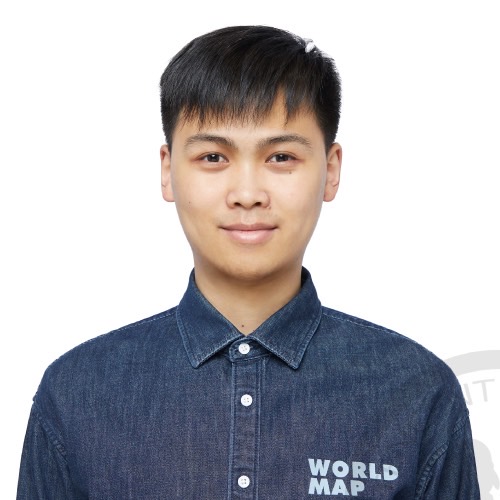}}]{Laixin Xie} is currently pursuing his Ph.D. degree at the School of Information Science and Technology, ShanghaiTech University. He received his B.S. Degree from Wuhan University of Science and Technology. His research interests include Explainable AI and Human-in-the-loop ML.
\end{IEEEbiography}
\vspace{-12mm}

\begin{IEEEbiography}[{\includegraphics[width=1in,height=1.25in,clip,keepaspectratio]{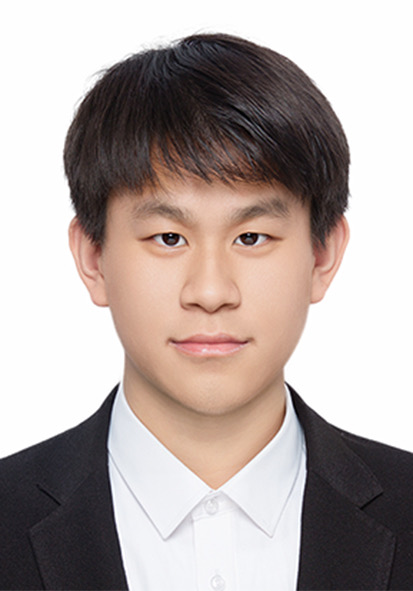}}]{Yang Ouyang} is currently  pursuing his Ph.D. degree at the School of Information Science and Technology, ShanghaiTech University. He received his B.S. degree from ShanghaiTech University. His research interests include high‑dimensional data visualization, eXplainable Artificial Intelligence (XAI), and visual analytics.
\end{IEEEbiography}
\vspace{-12mm}


\begin{IEEEbiography}[{\includegraphics[width=1in,height=1.25in,clip,keepaspectratio]{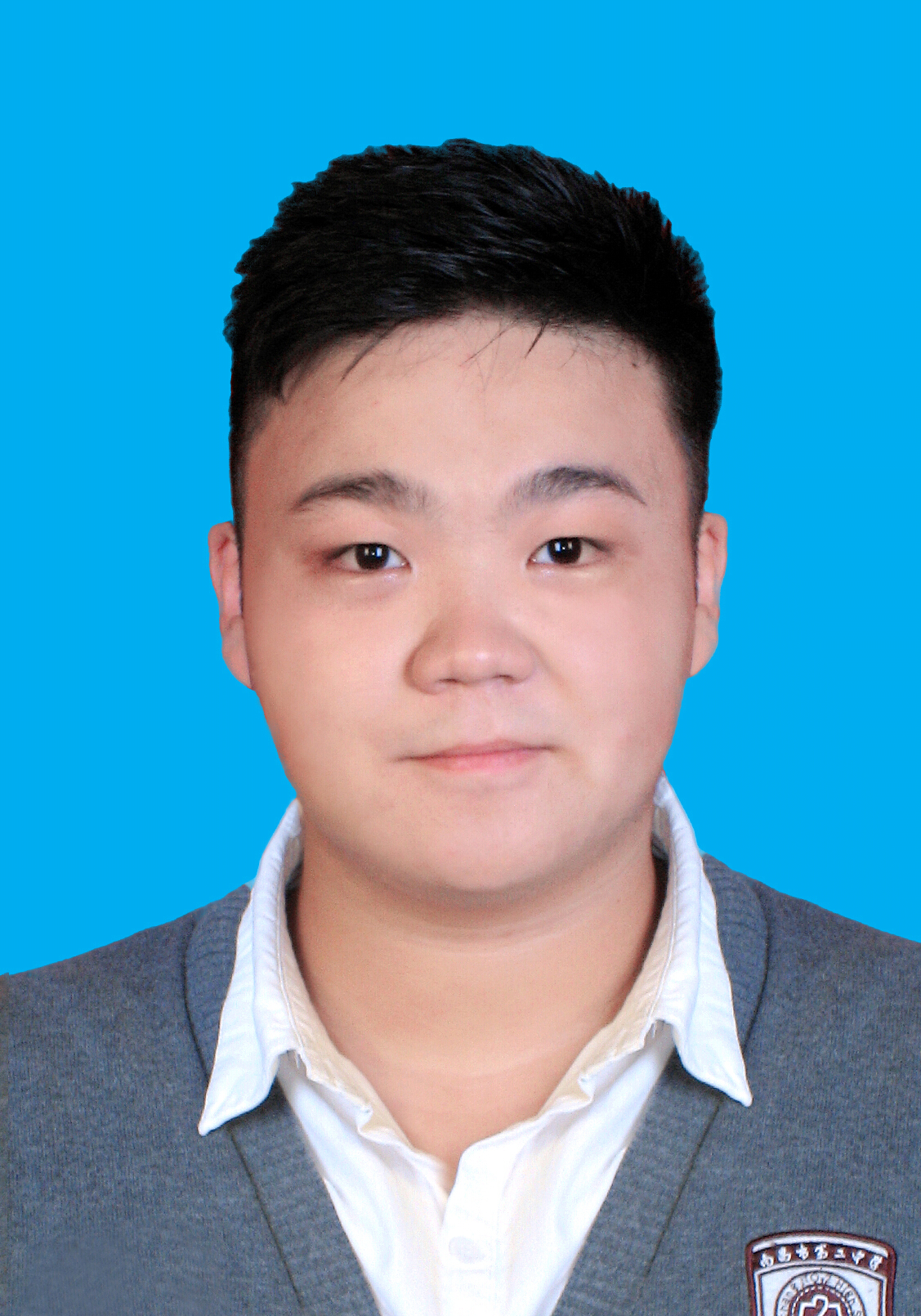}}]{Longfei Chen} is currently pursuing his M.S. degree at the School of Information Science and Technology, ShanghaiTech University. He received his B.S. degree from ShanghaiTech University. His research interests include computational social science and Explainable AI.
\end{IEEEbiography}
\vspace{-12mm}

\begin{IEEEbiography}[{\includegraphics[width=1in,height=1.25in,clip,keepaspectratio]{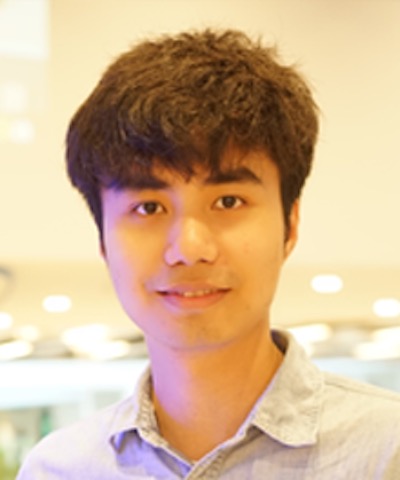}}]{Ziming Wu} is currently a senior researcher at Tencent, China. He obtained his Ph.D. degree in Computer Science and Technology at the Hong Kong University of Science and Technology. His research interests include large-scale social recommendation, computational social science and human computer interaction. He has published over $20$ academic papers in top-tier conferences and journals. He also regularly serves as an external reviewer in relevant conferences, including AAAI, SIGCHI, WebConf, KDD, etc.
\end{IEEEbiography}
\vspace{-12mm}

\begin{IEEEbiography}[{\includegraphics[width=1in,height=1.25in,clip,keepaspectratio]{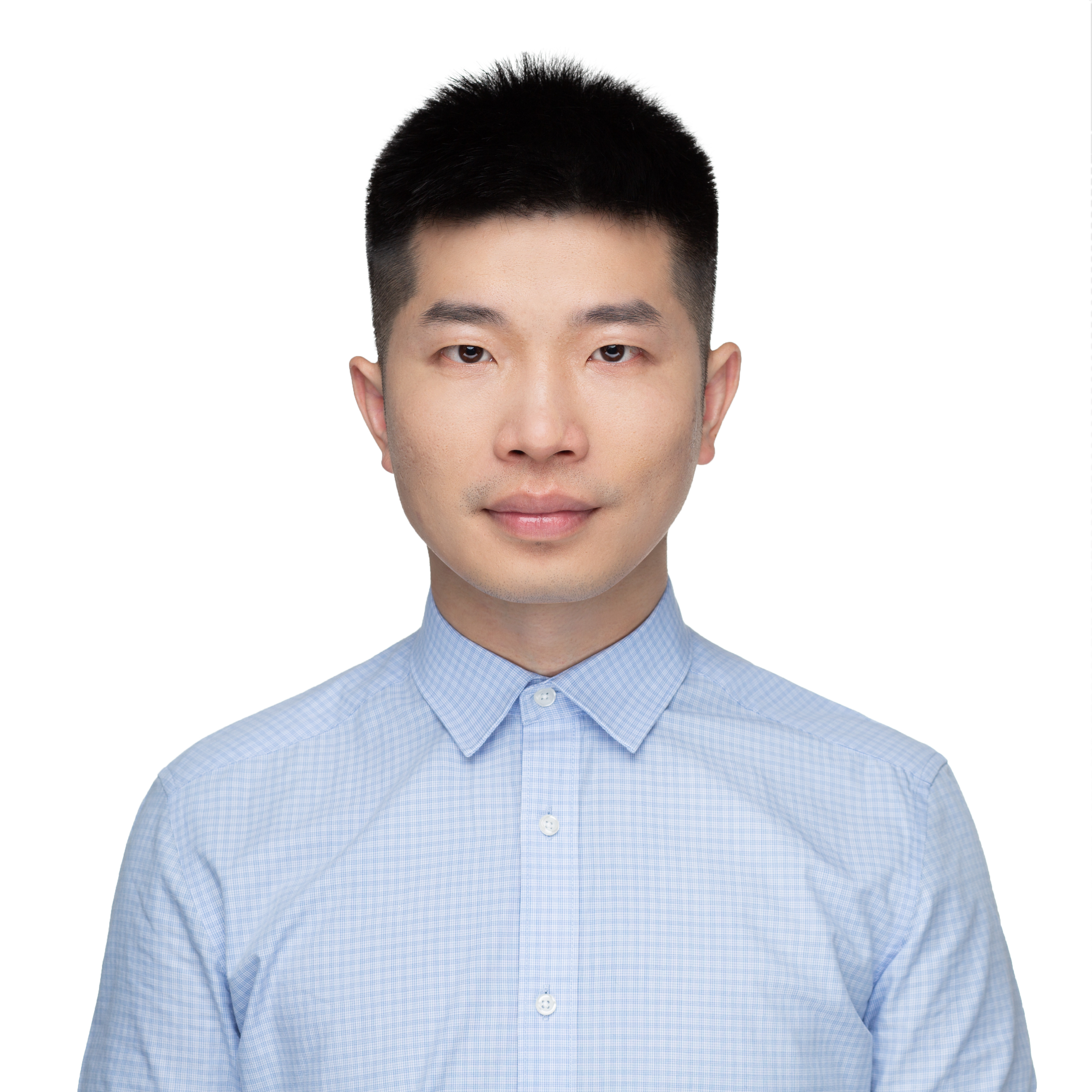}}]{Quan Li} is currently a tenure-track assistant professor at School of Information Science and Technology, ShanghaiTech University. He received his Ph.D. from the Department of Computer Science and Engineering at the Hong Kong University of Science and Technology. His research interests lie in artificial intelligence, big data visualization, and visual analysis, interpretable machine learning, data science, data storytelling, human-computer interaction (HCI), and computational social sciences. His academic achievements have been published in top journals and conferences on visualization and human-computer interaction. For more details, please refer to https://faculty.sist.shanghaitech.edu.cn/liquan/.
\end{IEEEbiography}




\end{document}